\documentclass[12pt]{article}%
\usepackage{amsmath}
\usepackage{amsfonts}
\usepackage{amssymb}
\usepackage{graphicx}
\usepackage{algorithmic}
\usepackage{algorithm}
\usepackage{hyperref}%
\providecommand{\U}[1]{\protect\rule{.1in}{.1in}}
\providecommand{\U}[1]{\protect\rule{.1in}{.1in}}
\newtheorem{theorem}{Theorem}

\newtheorem{proposition}[theorem]{Proposition}
\newtheorem{corollary}[theorem]{Corollary}

\begin{document}

\title{An Imprecise SHAP as a Tool for Explaining the Class Probability Distributions
under Limited Training Data}
\author{Lev V. Utkin, Andrei V. Konstantinov, and Kirill A. Vishniakov\\Peter the Great St.Petersburg Polytechnic University\\St.Petersburg, Russia\\e-mail: lev.utkin@gmail.com, andrue.konst@gmail.com, ki.vishniakov@gmail.com}
\date{}
\maketitle

\begin{abstract}
One of the most popular methods of the machine learning prediction explanation
is the SHapley Additive exPlanations method (SHAP). An imprecise SHAP as a
modification of the original SHAP is proposed for cases when the class
probability distributions are imprecise and represented by sets of
distributions. The first idea behind the imprecise SHAP is a new approach for
computing the marginal contribution of a feature, which fulfils the important
efficiency property of Shapley values. The second idea is an attempt to
consider a general approach to calculating and reducing interval-valued
Shapley values, which is similar to the idea of reachable probability
intervals in the imprecise probability theory. A simple special implementation
of the general approach in the form of linear optimization problems is
proposed, which is based on using the Kolmogorov-Smirnov distance and
imprecise contamination models. Numerical examples with synthetic and real
data illustrate the imprecise SHAP.


\end{abstract}

\section{Introduction}

An importance of the machine learning models in many applications and their
success in solving many applied problems lead to another problem which may be
an obstacle for using the models in areas where their prediction accuracy as
well as understanding is crucial, for example, in medicine, reliability
analysis, security, etc. This obstacle takes place for complex models which
can be viewed as black boxes because their users do not know how the models
act and are functioning. Moreover, the training process is also often unknown.
A natural way for overcoming the obstacle is to use a meta-model which could
explain the provided predictions. The explanation means that we have to select
features of an analyzed example which are responsible for the prediction of
the obtained black-box model or significantly impact on the corresponding
prediction. By considering the explanation of a single example, we say about
the so-called local explanation methods. They aim to explain the black-box
model locally around the considered example. Another explanation methods try
to explain predictions taking into account the whole dataset or its certain
part. The need of explaining the black-box models using the local or global
explanations in many applications motivated developing a huge number of
explanation methods which are described in detail in several comprehensive
survey papers
\cite{Belle-Papantonis-2020,Guidotti-2019,Liang-etal-2021,Marcinkevics-Vogt-20,Molnar-2019,Rudin-etal-21,Xie-Ras-etal-2020,Zablocki-etal-21,Zhang-Tino-etal-2020}%
.

Among all explanation methods, we select two very popular methods: the Local
Interpretable Model-Agnostic Explanation (LIME) \cite{Ribeiro-etal-2016} and
SHapley Additive exPlanations (SHAP)
\cite{Lundberg-Lee-2017,Strumbel-Kononenko-2010}. The basic idea behind LIME
is to build an approximating linear model around the explained example. In
order to implement the approximation, many synthetic examples are generated in
the neighborhood of the explained example with weights depending on distances
from the explained example. The linear regression model is constructed by
using the generated examples such that its coefficients can be regarded as
quantitative representation of impacts of the corresponding features on the prediction.

SHAP is inspired by game-theoretic Shapley values \cite{Shapley-1953} which
can be interpreted as average expected marginal contributions over all
possible subsets (coalitions) of features to the black-box model prediction.
In spite of two important shortcomings of SHAP, including its computational
complexity depending on the number of features and some ambiguity of available
approaches for removing features from consideration \cite{Covert-etal-20},
SHAP is widely used in practice and can be viewed as the most promising and
theoretically justified explanation method which fulfils several nice
properties \cite{Lundberg-Lee-2017}. Nevertheless, another difficulty of using
SHAP is how to deal with predictions in the form of probability distributions
which arise in multi-class classification and in machine learning survival
analysis. The problem is that SHAP in each iteration calculates a difference
between two predictions defined by different subsets of features in the
explained example. Therefore, the question is how to define the difference
between the probability distributions. One of the simplest ways in
classification is to consider a class with the largest probability. However,
this way may lead to incorrect results when the probabilities are comparable.
Moreover, the same approach cannot be applied to explanation of the survival
model predictions which are in the form of survival functions. It should be
noted that Covert et al. \cite{Covert-etal-20a} justified that the well-known
Kullback-Leibler (KL) divergence \cite{Kullback-Leibler-1951} can be applied
to the global explanation. It is obvious that the KL divergence can be applied
also to the local explanation. Moreover, the KL divergence can be replaced
with different distances, for example, $\chi^{2}$-divergence
\cite{Pearson-1900}, the relative J-divergence \cite{Dragomir-etal-2001},
Csiszar's $f$-divergence measure \cite{Csiszar-1967}. However, the use of all
these measures leads to such modifications of SHAP that obtained Shapley
values do not satisfy properties of original Shapley values. Therefore, one of
the problems for solving is to define a meaningful distance between two
predictions represented in the form of the class probability distributions,
which fulfils the Shapley value properties.

It should be noted that the probability distribution of classes may be
imprecise due to a limited number of training data. On the one hand, it can be
said that the class probabilities in many cases are not real probabilities as
measures defined on events, i.e., they are not relative frequencies of
occurrence of events. For example, the softmax function in neural networks
produce numbers (weights of classes), which have some properties of
probabilities, but not original probabilities. On the other hand,
probabilities of classes in random forests and in random survival forests can
be viewed as relative frequencies (probabilities are computed by counting the
percentage of different classes of examples at each leaf node), and they are
true probabilities. In both the cases, it is obvious that accuracy of the
class probability distributions depends on the machine learning model
predicting the distributions and on the amount of training data. It is
difficult to impact on improvement of the post-hoc machine learning model
representing as a black box, but we can take into account the imprecision of
probability distributions due to the lack of sufficient training data.

The considered imprecision can be referred to the epistemic uncertainty
\cite{Senge-etal-14} which represents our ignorance about a model caused by
the lack of observation data. Epistemic uncertainty can be resolved by
observing more data. There are several approaches to formalize the
imprecision. One of them is to consider a set of probability distributions
instead of the single one. Sets of distributions are produced by imprecise
statistical models, for example, by the linear-vacuous mixture or imprecise
$\varepsilon$-contamination model \cite{Walley91}, by the imprecise Dirichlet
model \cite{Walley96a}, by the constant odds-ratio model \cite{Walley91}, by
Kolmogorov--Smirnov bounds \cite{Johnson-Leone64}. It is obvious that the
above imprecise statistical models produce interval-valued probabilities of
events or classes, and the obtained interval can be regarded as a measure of
observation data insufficiency.

It turns out that the imprecision of the class probability distribution leads
to imprecision of Shapley values when SHAP is used for explaining the machine
learning model predictions. In other words, Shapley values become
interval-valued. This allows us to construct a new framework of the imprecise
SHAP or imprecise Shapley values, which provides a real accounting of the
prediction uncertainty. Having the interval-valued Shapley values, we can
compare intervals in order to select the most important features. It is
interesting to point out that properties of the efficiency and the linearity
of Shapley values play a crucial role in constructing basic rules of the
framework of imprecise Shapley values.

In summary, the following contributions are made in this paper:

1. A new approach for computing the marginal contribution of the $i$-th
feature in SHAP for interpretation of the class probability distributions as
predictions of a black-box machine learning model is proposed. It is based on
considering a distance between two probability distributions. Moreover,
Shapley values using the proposed approach fulfils the efficiency property
which is very important in correct explanation by means of SHAP.

2. An imprecise SHAP is proposed for cases when the class probability
distributions are imprecise, i.e., they are represented by convex sets of
distributions. The outcome of the imprecise SHAP is a set of interval-valued
Shapley values. Basic tools for dealing with imprecise class probability
distributions and for computing interval-valued Shapley values are introduced.

3. Implementation of the imprecise SHAP by using the Kolmogorov-Smirnov
distance is proposed. Complex optimization problems for computing
interval-valued Shapley values are reduced to finite sets of simple linear
programming problems.

4. The imprecise SHAP is illustrated by means of numerical experiments with
synthetic and real data.

The code of the proposed algorithm can be found in https://github.com/LightnessOfBeing/ImpreciseSHAP

The paper is organized as follows. Related work is in Section 2. A brief
introduction to Shapley values and SHAP itself is given in Section 3
(Background). A new approach for computing the marginal contribution of each
feature in SHAP for multiclassification problems is provided in Section 4. The
imprecise SHAP is introduced in Section 5. Application of the
Kolmogorov-Smirnov distance to the imprecise SHAP to get simple calculations
of the interval-valued Shapley values is considered in Section 6. Numerical
experiments with synthetic data and real data are given in Section 7.
Concluding remarks can be found in Section 8.

\section{Related work}

\textbf{Local interpretation methods. }An increasing importance of machine
learning models and algorithms leads to development of new explanation methods
taking into account various peculiarities of applied problems. As a result,
many models of the local interpretation has been proposed. Success and
simplicity of LIME resulted in development of several its modifications, for
example, ALIME \cite{Shankaranarayana-Runje-2019}, Anchor LIME
\cite{Ribeiro-etal-2018}, LIME-Aleph \cite{Rabold-etal-2019}, GraphLIME
\cite{Huang-Yamada-etal-2020}, SurvLIME \cite{Kovalev-Utkin-Kasimov-20a}, etc.
A comprehensive analysis of LIME, including the study of its applicability to
different data types, for example, text and image data, is provided by Garreau
and Luxburg \cite{Garreau-Luxburg-2020}. The same analysis for tabular data is
proposed by Garreau and Luxburg \cite{Garreau-Luxburg-2020a}. An image version
of LIME with its thorough theoretical investigation is presented by Garreau
and Mardaoui \cite{Garreau-Mardaoui-21}. An interesting information-theoretic
justification of interpretation methods on the basis of the concept of
explainable empirical risk minimization is proposed by Jung \cite{Jung-20}.

In order to relax the linearity condition for the local interpretation models
like LIME and to adequately approximate a black-box model, several methods
based on using Generalized Additive Models (GAMs)
\cite{Hastie-Tibshirani-1990} were proposed
\cite{Chang-Tan-etal-2020,Lou-etal-12,Nori-etal-19,Zhang-Tan-Koch-etal-19}.
Another interesting class of models based on using a linear combination of
neural networks such that a single feature is fed to each network was proposed
by Agarwal et al. \cite{Agarwal-etal-20}. The impact of every feature on the
prediction in these models is determined by its corresponding shape function
obtained by each neural network. Following ideas behind these interpretation
models, Konstantinov and Utkin \cite{Konstantinov-Utkin-20b} proposed a
similar model, but an ensemble of gradient boosting machine is used instead of
neural networks in order to simplify the explanation model training process.

Another explanation method is SHAP
\cite{Lundberg-Lee-2017,Strumbel-Kononenko-2010}, which takes a game-theoretic
approach for optimizing a regression loss function based on Shapley values.
General questions of the computational efficiency of SHAP were investigated by
Van den Broeck et al. \cite{Broeck-etal-21}. Bowen and Ungar
\cite{Bowen-Ungar-20} proposed the generalized SHAP method (G-SHAP) which
allows us to compute the feature importance of any function of a model's
output. Rozemberczki and Sarkar \cite{Rozemberczki-Sarkar-21} presented an
approach to applying SHAP to ensemble models. The problem of explaining the
predictions of graph neural networks by using SHAP was considered by Yuan et
al. \cite{Yuan-Yu-20}. Frye et al. \cite{Frye-etal-2020} introduced the
so-called off- and on-manifold Shapley values for high-dimensional multi-type
data. Application of SHAP to explanation of recurrent neural networks was
studied in \cite{Bento-etal-20}. Begley et al. present a new approach to
explaining fairness in machine learning, based on the Shapley value paradigm.
Antwarg et al. \cite{Antwarg-etal-20} studied how to explain anomalies
detected by autoencoders using SHAP. The problem of explaining anomalies
detected by PCA is also considered by Takeishi \cite{Takeishi-19}. SHAP is
also applied to problems of explaining individual predictions when features
are dependent \cite{Aas-etal-2019} or when features are mixed
\cite{Redelmeier-etal-20}. SHAP has been used in real applications to explain
predictions of the black-box models, for example, it was used to rank failure
modes of reinforced concrete columns and to explains why a machine learning
model predicts a specific failure mode for a given sample
\cite{Mangalathu-etal-20}. It was also used in chemoinformatics and medicinal
chemistry \cite{Rodriguez-20}.

A lot of interpretation methods, their analysis, and critical review can be
found also in survey papers
\cite{Adadi-Berrada-2018,Arrieta-etal-2019,Belle-Papantonis-2020,Carvalho-etal-2019,Das-Rad-20,Guidotti-2019,Liang-etal-2021,Rudin-2019,Xie-Ras-etal-2020}%
.

\textbf{Imprecise probabilities in classification and regression}. One of the
first ideas of applying imprecise probability theory to classification
decision trees was presented in \cite{Abellan-Moral-2003}, where probabilities
of classes at decision tree leaves are estimated by using an imprecise model,
and the so-called Credal Decision Tree model is proposed. Following this work,
several papers devoted to applications of imprecise probabilities to decision
trees and random forests were presented
\cite{Abellan-etal-2017,Abellan-etal-2018,Mantas-Abellan-2014,Moral-Garcia-etal-2020}%
, where the authors developed new splitting criteria taking into account
imprecision of training data and noisy data. In particular, the authors
consider the application of Walley's imprecise Dirichlet model (IDM)
\cite{Walley96a}. The main advantage of the IDM in its application to the
classification problems is that it produces a convex set of probability
distributions, which has nice properties and depends on a number of
observations. Another interesting model called the fuzzy random forest is
proposed in \cite{Bonissone-etal-2010}. As an alternative to the use of the
IDM, nonparametric predictive inference has also been used successfully for
imprecise probabilistic inference with decision trees \cite{Abellan-etal-2014}%
. Imprecise probabilities have also been used in classification problems in
\cite{Destercke-Antoine-2013,Matt-2017,Moral-2019}. The main focus of interest
in this paper is not imprecise probabilities in machine learning models, but
imprecision of Shapley values as a consequence of the machine learning model
prediction imprecision when SHAP is used to explain the model prediction.

\section{Shapley values and model explainability}

One of the approaches to explaining machine learning model predictions is the
Shapley value \cite{Shapley-1953} as a concept in coalitional games. According
to the concept, the total gain of a game is distributed among players such
that desirable properties, including efficiency, symmetry, and linearity, are
fulfilled. In the framework of the machine learning, the gain can be viewed as
the machine learning model prediction or the model output, and a player is a
feature of input data. Hence, contributions of features to the model
prediction can be estimated by Shapley values. The $i$-th feature importance
is defined by the Shapley value%
\begin{equation}
\phi_{i}(f)=\phi_{i}=\sum_{S\subseteq N\backslash\{i\}}B(S,N)\left[  f\left(
S\cup\{i\}\right)  -f\left(  S\right)  \right]  , \label{SHAP_1}%
\end{equation}
where $f\left(  S\right)  $ is the black-box model prediction under condition
that a subset $S$ of features are used as the corresponding input; $N$ is the
set of all features; $B(S,N)$ is defined as
\begin{equation}
B(S,N)=\frac{\left\vert S\right\vert !\left(  \left\vert N\right\vert
-\left\vert S\right\vert -1\right)  !}{\left\vert N\right\vert !}.
\end{equation}

It can be seen from (\ref{SHAP_1}) that the Shapley value $\phi_{i}$ can be
regarded as the average contribution of the $i$-th feature across all possible
permutations of the feature set.

The Shapley value has the following well-known properties:

\textbf{Efficiency}. The total gain is distributed as $\sum_{k=0}^{m}\phi
_{k}=f(\mathbf{x})-f(\varnothing).$

\textbf{Symmetry}. If two players with numbers $i$ and $j$ make equal
contributions, i.e., $f\left(  S\cup\{i\}\right)  =f\left(  S\cup\{j\}\right)
$ for all subsets $S$ which contain neither $i$ nor $j$, then $\phi_{i}%
=\phi_{j}$.

\textbf{Dummy}. If a player makes zero contribution, i.e., $f\left(
S\cup\{j\}\right)  =f\left(  S\right)  $ for a player $j$ and all $S\subseteq
N\backslash\{j\}$, then $\phi_{j}=0$.

\textbf{Linearity}. A linear combination of multiple games $f_{1},...,f_{n}$,
represented as $f(S)=\sum_{k=1}^{n}c_{k}f_{k}(S)$, has gains derived from $f$:
$\phi_{i}(f)=\sum_{k=1}^{m}c_{k}\phi_{i}(f_{k})$ for every $i$.

Let us consider a machine learning problem. Suppose that there is a dataset
$\{(\mathbf{x}_{1},y_{1}),...,(\mathbf{x}_{n},y_{n})\}$ of $n$ points
$(\mathbf{x}_{i},y_{i})$, where $\mathbf{x}_{i}\in\mathcal{X}\subset
\mathbb{R}^{m}$ is a feature vector consisting of $m$ features, $y_{i}$ is the
observed output for the feature vector $\mathbf{x}_{i}$ such that $y_{i}%
\in\mathbb{R}$ in the regression problem and $y_{i}\in\{1,2,...,T\}$ in the
classification problem with $T$ classes.

If the task is to interpret or to explain prediction $f(\mathbf{x}^{\ast})$
from the model at a local feature vector $\mathbf{x}^{\ast}$, then the
prediction $f(\mathbf{x}^{\ast})$ can be represented by using Shapley values
as follows \cite{Lundberg-Lee-2017,Strumbel-Kononenko-2010}:
\begin{equation}
f(\mathbf{x}^{\ast})=\phi_{0}+\sum_{j=1}^{m}\phi_{j}^{\ast},
\end{equation}
where $\phi_{0}=\mathbb{E}[f(\mathbf{x})]$, $\phi_{j}^{\ast}$ is the value
$\phi_{j}$ for the prediction $\mathbf{x}=\mathbf{x}^{\ast}$.

The above implies that Shapley values explain the difference between
prediction $f(\mathbf{x}^{\ast})$ and the global average prediction.

One of the crucial questions for implementing SHAP is how to remove features
from subset $N\backslash S$, i.e., how to fill input features from subset
$N\backslash S$ in order to get predictions $f\left(  S\right)  $ of the
black-box model. A detailed list of various ways for removing features is
presented by Covert at al. \cite{Covert-etal-20}. One of the ways is simply by
setting the removed features to zero
\cite{Petsiuk-etal-2018,Zeiler-Fergus-2014} or by setting them to user-defined
default values \cite{Ribeiro-etal-2016}. According to the way, features are
often replaced with their mean values. Another way removes features by
replacing them with a sample from a conditional generative model
\cite{Yu-Lin-Yang-etal-18}. In the LIME method for tabular data, features are
replaced with independent draws from specific distributions
\cite{Covert-etal-20} such that each distribution depends on original feature
values. These are only a part of all ways for removing features.

\section{SHAP and multiclassification problems}

\subsection{SHAP and the Kullback-Leibler divergenc}

Let us consider a classification model $f$ whose predictions for the subset
$S\cup\{i\}$ are probabilities $p_{1},...,p_{C}$ of $C$ classes under
conditions $p_{1}+...+p_{C}=1$ and $p_{k}\geq0$ for all $k$. The probabilities
can be regarded as a distribution $P(S)=(p_{1},...,p_{C})$ over a $C$-class
categorical variable. Suppose that the prediction of the model $f$ for the
subset $S$ is a distribution denoted as $Q(S)=(q_{1},...,q_{C})$, where
$q_{1}+...+q_{C}=1$ and $q_{k}\geq0$ for all $k$.

Covert et al. \cite{Covert-etal-20a} justified that the Kullback-Leibler
divergence $KL(P||Q)$ is a natural way to measure the deviation of the
predictions from $P$ and $Q$ for the global explanation. Indeed, this is a
very interesting idea to consider the KL divergence instead of the difference
$f\left(  S\cup\{i\}\right)  -f\left(  S\right)  $. Lower the KL divergence
value, the better we have matched the true distribution with our
approximation, and the smaller we have impact of the $i$-th feature. Then
(\ref{SHAP_1}) can be rewritten as%

\begin{equation}
\phi_{i}(f)=\phi_{i}=\sum_{S\subseteq N\backslash\{i\}}B(S,N)KL\left(
P(S)||Q(S)\right)  .
\end{equation}

Let us return to some properties of the Kullback-Leibler divergence. First, it
is assumed that either $q_{i}\neq0$ for all values of $i$, or that if one
$p_{i}=0$, then $q_{i}=0$ as well. In this case, there holds $0/0=1$. Second,
it is called the information gain achieved if $P$ would be used instead of $Q$
which is currently used. In Bayesian interpretation, the KL divergence shows
updating from a prior distribution $Q$ (without knowledge the $i$-th feature)
to the posterior distribution $P$ (with the available $i$-th feature). Third,
the KL divergence is a non-symmetric measure. This idea has a deeper meaning.
It means that generally $KL(P||Q)\neq KL(Q||P)$. The last property is very
important. Indeed, when we say that the $i$-th feature positively contributes
into the prediction for some $S$, this means that $f\left(  S\cup\{i\}\right)
-f\left(  S\right)  >0$ and does not mean that $f\left(  S\right)  -f\left(
S\cup\{i\}\right)  >0$, i.e., the contribution is also non-symmetric to some
extent. This implies that the replacement of the difference $f\left(
S\cup\{i\}\right)  -f\left(  S\right)  $ with measure $KL$ should fulfil the
following property: $KL(P||Q)\geq KL(Q||P)$ if the $i$-th feature positively
contributes into the prediction. This also means that we cannot use one of the
standard symmetric distance metrics for getting the difference between
predictions. It should distinguish here the symmetry property of the
divergence measure and the symmetry property of the Shapley values.

\subsection{Interpretation of the predicted distribution $P_{S}$}

Suppose that the black-box classification model provides a class probability
distribution $P_{N}=f(N)$ as its prediction for the input data with features
from the set $N$. Without loss of generality, we assume that $P_{N}$ is close
to the vector $(1,0,...,0)$. In order to use the SHAP, we have to consider two
types of probability distributions: $P_{S\cup i}=f\left(  S\cup\{i\}\right)  $
and $P_{S}=f\left(  S\right)  $, where $S\subseteq N\backslash\{i\}$ (see
(\ref{SHAP_1})).

\begin{figure}
[ptb]
\begin{center}
\includegraphics[
height=1.7548in,
width=4.6727in
]%
{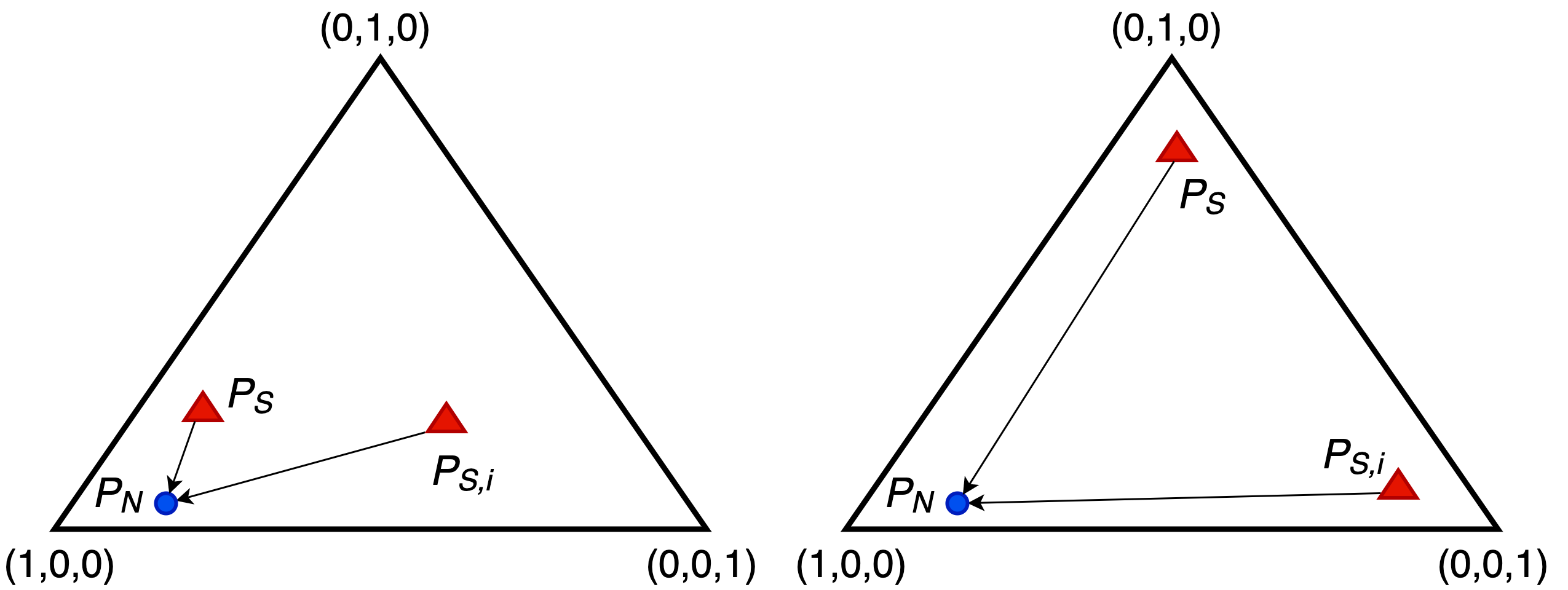}%
\caption{Two cases of the relationship between probability distributions
$P_{N}$, $P_{S}$, $P_{S,i}$}%
\label{fig:SHAP_simplex_1}%
\end{center}
\end{figure}

Let us define a function of predictions $P_{S}$ as well as $P_{S,i}$ for all
$S$ and $i$ as the distance between $P_{N}$ and $P_{S}$ or $P_{N}$ and
$P_{S\cup i}$, denoted as $D(P_{S},P_{N})$ or $D(P_{S,i},P_{N})$,
respectively. Let us also introduce the marginal contribution of the $i$-th
feature as $D(P_{S},P_{N})-D(P_{S,i},P_{N})$. This is an unusual definition of
the marginal contribution can be explained by means of Fig.
\ref{fig:SHAP_simplex_1} where two cases of the predicted probability
distributions of examples having three features are illustrated in the
probabilistic unit simplex with vertices $(1,0,0)$, $(0,1,0)$, $(0,0,1)$.
Distributions $P_{S\cup i}$ and $P_{S}$ are depicted by small triangles.
Distribution $P_{N}$ is depicted by the small circle.

At first glance, it seems that the contribution of the $i$-th feature should
be measured by the distance between points $P_{S}$ and $P_{S,i}$. In
particular, if the points coincide, then the distance is zero, and there is no
contribution of the considered feature into the prediction or into the class
probability distribution. Moreover, Covert et al. \cite{Covert-etal-20a}
proposed to apply the KL divergence $KL\left(  P_{S,i}||P_{S}\right)  $ to
characterize the feature marginal contribution. However, if we look at the
right unit simplex in Fig. \ref{fig:SHAP_simplex_1}, then we can see that the
distance between $P_{S}$ and $P_{S,i}$ is very large. Hence, the marginal
contribution of the $i$-th feature should be large. However, after adding the
$i$-th feature, we remain to be at the same distance from $P_{N}$. We do not
improve our knowledge about the true class or about $P_{N}$ after adding the
$i$-th feature to $S$ and computing the corresponding prediction. Note that we
do not search for the contribution of the $i$-th feature into the prediction
$P_{S}$. We aim to estimate how the $i$-th feature contributes into $P_{N}$.
Points $P_{S}$ and $P_{S,i}$ in the right simplex are approximately at the
same distance from $P_{N}$, therefore, the difference $D(P_{S},P_{N}%
)-D(P_{S,i},P_{N})$ is close to $0$ though points $P_{S}$ and $P_{S,i}$ are
far from each other. The left unit simplex in Fig. \ref{fig:SHAP_simplex_1}
shows a different case when $P_{S,i}$ negatively contributes into our
knowledge about the class distribution $P_{N}$ because it makes the
probability distribution to be more uncertain in comparison with $P_{S}$. This
implies that the difference $D(P_{S},P_{N})-D(P_{S,i},P_{N})$ should be
negative in this case.

Assuming that $D(P_{S},P_{N})$ is the KL divergence, this distance can be
regarded as the information gain achieved if $P_{S}$ would be used instead of
$P_{N}$. A similar information gain is achieved if $P_{S,i}$ would be used
instead of $P_{N}$. Then we can conclude that the difference $D(P_{S}%
,P_{N})-D(P_{S,i},P_{N})$ can be regarded as the contribution of the $i$-th
feature in the change of information gain.%

\begin{figure}
[ptb]
\begin{center}
\includegraphics[
height=1.8909in,
width=2.4989in
]%
{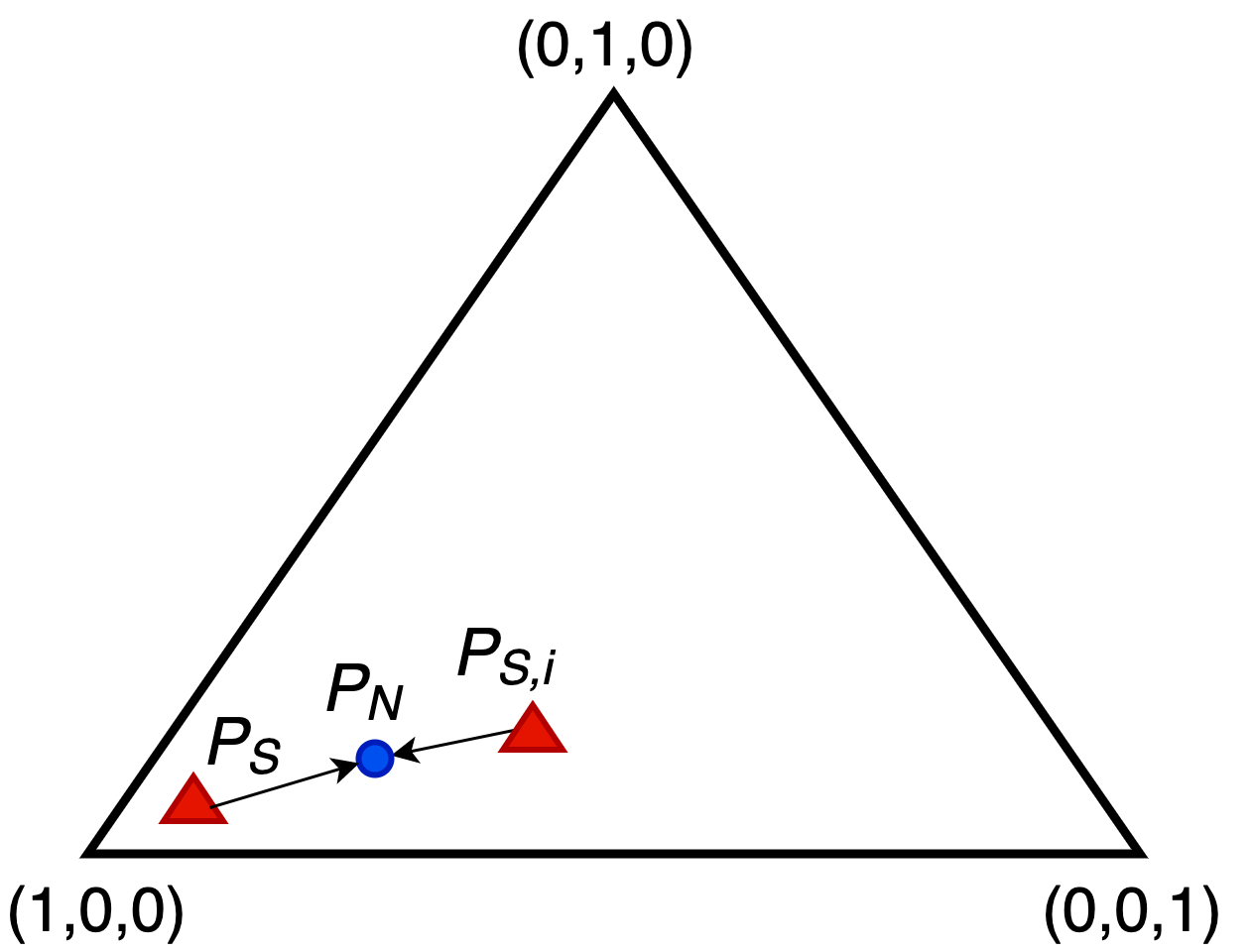}%
\caption{A specific case of the probability distribution location}%
\label{fig:SHAP_simplex_2}%
\end{center}
\end{figure}

The problem arises when we consider a case illustrated in Fig.
\ref{fig:SHAP_simplex_2}, where points $P_{S}$ and $P_{S,i}$ are at the same
distance from $P_{N}$. However, this case can be also explained in the same
way. We do not interpret a final decision about a class of the example. The
distribution $P_{0}$ is interpreted here. But this distribution is correctly
interpreted because the distribution $P_{S,i}$ does not contribute into
$P_{N}$ in comparison with the distribution $P_{S}$. The distribution $P_{S}$
explains the decision $(1,0,0)$ better than $P_{S,i}$, but not than $P_{N}$.
Here we meet a situation when interpretation strongly depends on its aim. If
its aim is to interpret $P_{N}$ without taking into account the final decision
about a class, then we should use the above approach, and contribution of the
$i$-th feature in the case illustrated in Fig. \ref{fig:SHAP_simplex_2} is $0$.%

\subsection{Interpretation of the predicted class}

\begin{figure}
[ptb]
\begin{center}
\includegraphics[
height=1.9386in,
width=2.5467in
]%
{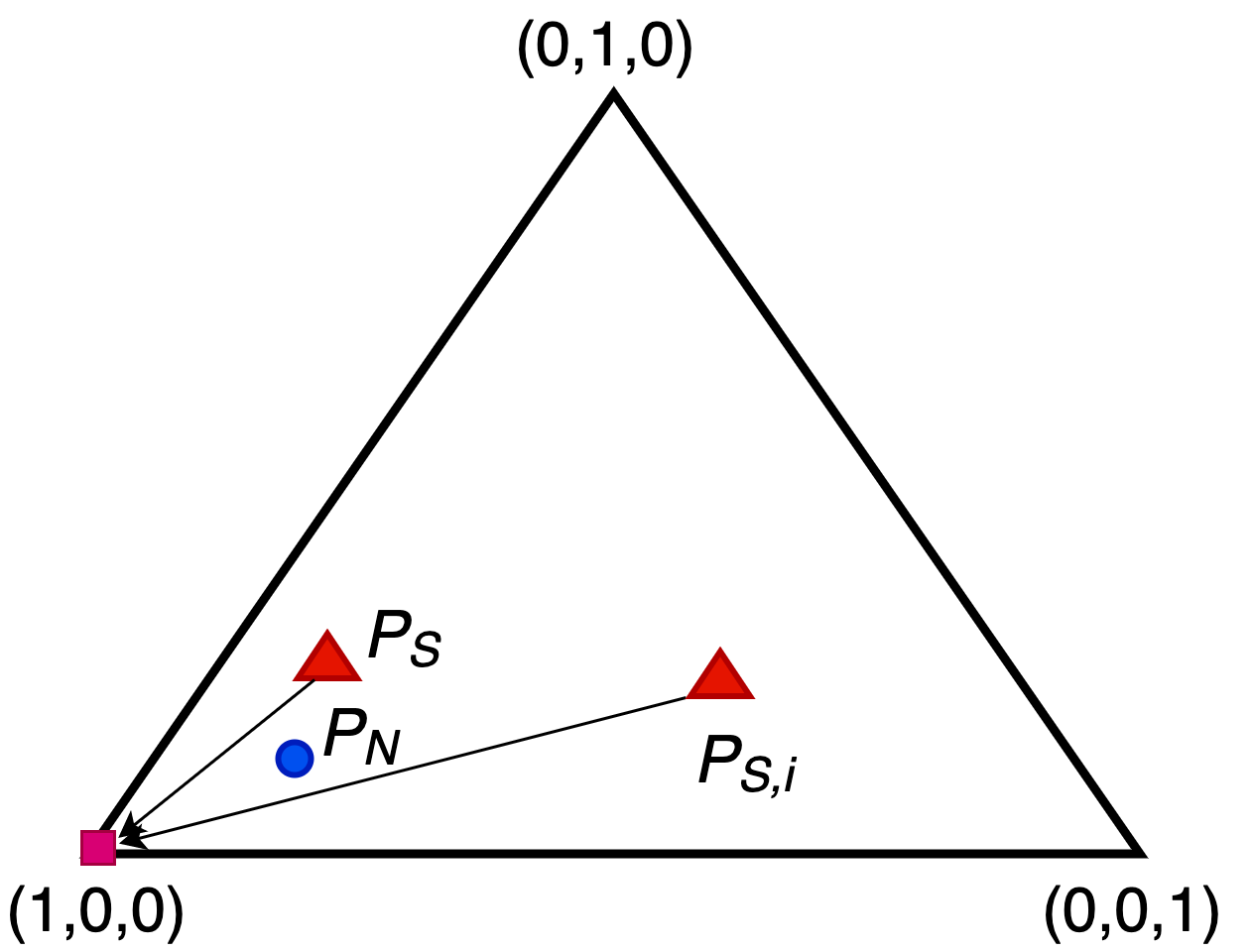}%
\caption{Interpretation of the predicted class}%
\label{fig:SHAP_simplex_3}%
\end{center}
\end{figure}

Another case is when our aim is to explain the final decision about the
predicted class by having predictions in the form of probability
distributions. The class can be represented by the probability distribution
$P_{0}=(1,0,...,0)$ which is depicted in Fig. \ref{fig:SHAP_simplex_3} by a
small square. In this case, we compare distributions $P_{S}$ and $P_{S,i}$
with $P_{0}$, but not with $P_{N}$ because we analyze how the $i$-th feature
contributes into change of the decision about the class of the corresponding
example, or how this feature contributes into our knowledge about the true
class. This is an important difference from the above case when we interpreted
$P_{N}$. It can be seen from Fig. \ref{fig:SHAP_simplex_3} that we compute
distances $D(P_{S},P_{0})$ and $D(P_{S,i},P_{0})$ instead of $D(P_{S},P_{N})$
and $D(P_{S,i},P_{N})$, respectively. When interpretation of the class is
performed, the contribution of the $i$-th feature for the case shown in Fig.
\ref{fig:SHAP_simplex_2} is no longer zero. Therefore, we will distinguish the
above two cases: interpretation of the class probability distribution and
interpretation of the predicted class by using the predicted class probability distribution.%

It seems that\ the distance $D(P_{S},P_{0})$\ could be viewed as a measure of
uncertainty of the prediction $P_{S}$. However, if we assume that
$P_{0}=(1,0,0)$ and $P_{S}=(0,1,0)$, then the distance between distributions
is largest, but the distribution $P_{S}$ is certain. In order to consider the
uncertainty measure, the distance $D(P_{S},P_{u})$ should be studied in place
of $D(P_{S},P_{0})$, where $P_{u}$ is the uniform distribution $(1/C,...,1/C)$%
. In this case, the distance $D(P_{S},P_{u})$ is a certainty measure of
$P_{S}$, because the most uncertain case when $P_{S}=P_{u}$ leads to
$D(P_{S},P_{u})=0$.

\subsection{Properties of Shapley values and the introduced distances}

In sum, the Shapley values are computed now as follows:
\begin{equation}
\phi_{i}=\sum_{S\subseteq N\backslash\{i\}}B(S,N)\left[  D(P_{S}%
,P_{N})-D(P_{S,i},P_{N})\right]  . \label{SHAP_prob_10}%
\end{equation}

Let us study how the introduced distances fulfil basic properties of Shapley
values. Their proof directly follows from the representation of the prediction
$f(S)$ as $-D(P_{S},P_{N})$ or $-D(P_{S},P_{0})$. Since the condition
$f(\emptyset)=0$ may be not valid (the prediction by some predefined feature
values), then the efficiency property is rewritten as follows. For the case of
the interpretation of the predicted distribution $P_{S}$, the total gain is
determined as
\begin{align}
\sum_{k=1}^{m}\phi_{k}  &  =f(\mathbf{x})-f(\emptyset)=-D(P_{N},P_{N}%
)+D(P_{\emptyset},P_{N})\nonumber\\
&  =D(P_{\emptyset},P_{N}). \label{SHAP_prob_11}%
\end{align}
Here $P_{\emptyset}$ is the black-box model prediction by replacing all
features with some predefined values, for example, with mean values.

Other properties of the Shapley values, including the symmetry, dummy and
linearity properties, remain without changes.

\section{Imprecise extension of SHAP}

\subsection{SHAP by sets of probability distributions (a general approach)}

We assume that a prediction of the black-box model is imprecise due to due to
a small amount of training data, due to our ignorance about a model caused by
the lack of observation data. In order to take into account the imprecision,
it is proposed to replace precise probability distributions of classes by sets
of probability distributions denoted as $\mathcal{P}$, which can be
constructed in accordance with one of the well-known imprecise statistical
inference models \cite{Walley91}.

The imprecision of predictions changes the definition of distances between
probability distributions because precise distributions are replaced with sets
of probability distributions $\mathcal{P}$. This implies that distances
between sets $\mathcal{P}_{1}$ and $\mathcal{P}_{2}$ of distributions should
be defined instead of distances between single distributions. We have now a
set of distances between all pairs of points such that one point in each pair
belongs to set $\mathcal{P}_{1}$, and another point in the pair belongs to set
$\mathcal{P}_{2}$. As a results, the set of distances produces an interval
with some lower and upper bounds corresponding to the smallest and the largest
distances, respectively. The same can be said about pairs of distributions
which produce an interval of Shapley values $\phi_{i}$ denoted as $[\phi
_{i}^{L},\phi_{i}^{U}]$. This implies that (\ref{SHAP_prob_10}) can be
rewritten as
\begin{equation}
\phi_{i}^{L}=\sum_{S\subseteq N\backslash\{i\}}B(S,N)\min_{P\in\mathcal{P}%
(P_{S}),~R\in\mathcal{P}(P_{S,i}),~Q\in\mathcal{P}(P_{N})}\left[
D(P,Q)-D(R,Q)\right]  , \label{SHAP_prob_21}%
\end{equation}%
\begin{equation}
\phi_{i}^{U}=\sum_{S\subseteq N\backslash\{i\}}B(S,N)\max_{P\in\mathcal{P}%
(P_{S}),~R\in\mathcal{P}(P_{S,i}),~Q\in\mathcal{P}(P_{N})}\left[
D(P,Q)-D(R,Q)\right]  , \label{SHAP_prob_22}%
\end{equation}
where $P$, $R$, $Q$ are probability distributions from subsets $\mathcal{P}%
(P_{S})$, $\mathcal{P}(P_{S,i})$, $\mathcal{P}(P_{N})$, respectively.

We do not define strongly subsets $\mathcal{P}(P_{S})$, $\mathcal{P}(P_{S,i}%
)$, $\mathcal{P}(P_{N})$ in order to derive general results. We only assume
that every subset is convex and is a part of the unit simplex of
probabilities. Specific imprecise statistical models producing sets of
probability distributions will be considered below.

The next question is how to calculate intervals for Shapley values. A simple
way is to separately compute the minimum and the maximum of distances $D$,
namely, as follows:
\begin{equation}
\phi_{i}^{L}=\sum_{S\subseteq N\backslash\{i\}}B(S,N)\left[  \min
_{P\in\mathcal{P}(P_{S}),~Q\in\mathcal{P}(P_{N})}D(P,Q)-\max_{R\in
\mathcal{P}(P_{S,i}),~Q\in\mathcal{P}(P_{N})}D(R,Q)\right]  ,
\end{equation}%
\begin{equation}
\phi_{i}^{U}=\sum_{S\subseteq N\backslash\{i\}}B(S,N)\left[  \max
_{P\in\mathcal{P}(P_{S}),~Q\in\mathcal{P}(P_{N})}D(P,Q)-\min_{R\in
\mathcal{P}(P_{S,i}),~Q\in\mathcal{P}(P_{N})}D(R,Q)\right]  ,
\end{equation}

However, obtained intervals of Shapley values may be too wide because we
actually consider extreme cases assuming that the distribution $Q\in
\mathcal{P}(P_{N})$ may be different in different distances, i.e., in $D(P,Q)$
and $D(R,Q)$. This assumption might be reasonable. Moreover, it would
significantly simplify the optimization problems. In order to obtain tighter
intervals, we return to (\ref{SHAP_prob_21})-(\ref{SHAP_prob_22}). We assume
that there exists a single class probability distribution $Q$ from
$\mathcal{P}(P_{N})$, which is unknown, but it provides the smallest or the
largest value of $D(P,Q)-D(R,Q)$ over all distributions from $\mathcal{P}%
(P_{N})$. Of course, we relax this condition in other terms of the sum over
$S\subseteq N\backslash\{i\}$ because it is extremely difficult to solve the
obtained optimization problem under this condition. The same can be said about
distributions $P$ and $R$. As a result, we also get wide intervals for Shapley
values because, but they will be reduced taking into account intervals for all
Shapley values and the efficiency property of the values.

Solutions of optimization problems in (\ref{SHAP_prob_21}) and
(\ref{SHAP_prob_22}) depend on definitions of the distance $D$. Let us return
to the efficiency property (\ref{SHAP_prob_11}) which is briefly written as
$\sum_{k=0}^{m}\phi_{k}=D(P_{\emptyset},P_{N})$. Suppose that we have computed
$\phi_{i}^{L}$ and $\phi_{i}^{U}$. Let us denote the lower and upper bounds
for $D(P_{\emptyset},P_{N})$ as $D^{L}$ and $D^{U}$, respectively. Then there
hold
\begin{equation}
D^{L}=\min_{T\in\mathcal{P}(P_{\emptyset}),\ Q\in\mathcal{P}(P_{N})}D(T,Q),
\end{equation}%
\begin{equation}
D^{U}=\max_{T\in\mathcal{P}(P_{\emptyset}),\ Q\in\mathcal{P}(P_{N})}D(T,Q).
\end{equation}

Hence, the efficiency property of Shapley values can be rewritten by taking
into account interval-valued Shapley values as
\begin{equation}
D^{L}\leq\sum_{k=1}^{m}\phi_{k}\leq D^{U}. \label{SHAP_prob_25}%
\end{equation}

It follows from (\ref{SHAP_prob_25}) that we cannot write a precise version of
the efficiency property because we do not know precise distributions
$P_{\emptyset}$ and $P_{N}$. Therefore, we use bounds for the total gain, but
these bounds can help us to reduce intervals of $\phi_{i}$. Reduced intervals
can be obtained by solving the following linear optimization problems:
\begin{equation}
\tilde{\phi}_{k}^{L}=\min\phi_{k},\ \ \ \tilde{\phi}_{k}^{U}=\max\phi_{k},
\label{SHAP_prob_26}%
\end{equation}
subject to (\ref{SHAP_prob_25}) and $\phi_{k}^{L}\leq\phi_{k}\leq\phi_{k}^{U}%
$, $k=1,...,m$.

The above optimization problems for all $k$ from $1$ to $m$ allow us to get
tighter bounds for Shapley values. It turns out that problems
(\ref{SHAP_prob_26}) can be explicitly solved. In order to avoid introducing
new notations, we assume that all Shapley values are positive. It can be done
by subtracting value $\min_{i=1,...,m}\phi_{i}^{L}$ from all variables
$\phi_{i}$.

\begin{proposition}
\label{prop:imp_SHAP_1}Assume that $\min_{i=1,...,m}\phi_{i}^{L}\geq0$ and
\begin{equation}
\sum_{i=1}^{m}\phi_{i}^{L}\leq D^{L},\ \sum_{i=1}^{m}\phi_{i}^{U}\geq D^{U}.
\label{SHAP_prob_27}%
\end{equation}
Then problems (\ref{SHAP_prob_26}) have the following solutions for all
$k\in\{1,...,m\}$:
\begin{equation}
\tilde{\phi}_{k}^{U}=\min\left(  \phi_{k}^{U},D^{U}-\sum_{i=1,i\neq k}^{m}%
\phi_{i}^{L}\right)  , \label{SHAP_prob_28}%
\end{equation}%
\begin{equation}
\tilde{\phi}_{k}^{L}=\max\left(  \phi_{k}^{L},D^{L}-\sum_{i=1,i\neq k}^{m}%
\phi_{i}^{U}\right)  . \label{SHAP_prob_29}%
\end{equation}

\end{proposition}

It is interesting to note that the above problem statement is similar to the
definitions of reachable probability intervals \cite{Destercke-Antoine-2013}
in the framework of the imprecise probability theory \cite{Walley91}. It can
be regarded as an extension of the probabilistic definitions which consider
sets of probability distributions as convex subsets of the unit simplex.
Moreover, vectors of Shapley values produce a convex set. In contrast to the
imprecise probabilities, imprecise Shapley values are not a part of the unit
simplex. Transferring definitions of reachable probability intervals to the
above problem, we can write that conditions (\ref{SHAP_prob_27}) imply that
intervals $[\phi_{i}^{L},\phi_{i}^{U}]$ are proper. This implies that the
corresponding set of Shapley values is not empty. Moreover, it can be simply
proved that bounds $\tilde{\phi}_{k}^{L}$ and $\tilde{\phi}_{k}^{U}$ are
reachable. If the reachability condition is not satisfied, then intervals
$[\phi_{i}^{L},\phi_{i}^{U}]$ are unnecessarily broad. Moreover, they might be
such that some values of the intervals do not correspond to condition
(\ref{SHAP_prob_25}). Proposition \ref{prop:imp_SHAP_1} gives a way to compute
the reachable intervals of Shapley values. At the same time, it can be
extended to a more general case when we would like to find bounds for a linear
function $g(\phi_{1},...,\phi_{m})=\sum_{i=1}^{m}$ $a_{i}\phi_{i}=\left\langle
\mathbf{a},\mathbf{\phi}\right\rangle $ of imprecise Shapley values. Here
$\mathbf{a=(}a_{1}....,a_{m})$ is a vector of known coefficients,
$\mathbf{\phi}=(\phi_{1},...,\phi_{m})$. Intervals of $g(\mathbf{\phi
}|\mathbf{a})$ are computed by solving the following two linear programming
problems (minimization and maximization):
\[
g^{L}(\mathbf{\phi}|\mathbf{a})(g^{U}(\mathbf{\phi}|\mathbf{a}))=\min
(\max)\left\langle \mathbf{a},\mathbf{\phi}\right\rangle ,
\]
subject to (\ref{SHAP_prob_25}) and $\phi_{k}^{L}\leq\phi_{k}\leq\phi_{k}^{U}%
$, $k=1,...,m$.

The dual optimization problems are
\[
g^{L}(\mathbf{\phi}|\mathbf{a})=\max\left(  D^{L}v_{0}-D^{U}w_{0}+\sum
_{i=1}^{m}\phi_{i}^{L}v_{i}-\sum_{i=1}^{m}\phi_{i}^{U}w_{i}\right)  ,
\]
subject to
\[
v_{0}-w_{0}+\sum_{i=1}^{m}\left(  v_{i}-w_{i}\right)  \leq a_{k},\ k=1,...,m,
\]
and
\[
g^{U}(\mathbf{\phi}|\mathbf{a})=\min\left(  D^{U}v_{0}-D^{L}w_{0}+\sum
_{i=1}^{m}\phi_{i}^{U}v_{i}-\sum_{i=1}^{m}\phi_{i}^{L}w_{i}\right)  ,
\]
subject to
\[
v_{0}-w_{0}+\sum_{i=1}^{m}\left(  v_{i}-w_{i}\right)  \geq a_{k},\ k=1,...,m.
\]

The above comments can be viewed as a starting point for development of a
general theory of imprecise Shapley values. Moreover, Proposition
\ref{prop:imp_SHAP_1} provides a tool for dealing with intervals of Shapley
values without assumptions about sets $\mathcal{P}$ of class probability
distributions and about the distance $D$ between the sets. Therefore, the next
questions are to define imprecise statistical models producing sets
$\mathcal{P}$ and the corresponding distances $D$ such that bounds
(\ref{SHAP_prob_21})-(\ref{SHAP_prob_22}) could be computed in a simple way.
In order to answer these questions, we have to select a model producing
$\mathcal{P}$ and to select a distance $D$.

\subsection{Imprecise statistical models}

There are several imprecise models of probability distributions. One of the
interesting models is the linear-vacuous mixture or the imprecise
$\varepsilon$-contamination model \cite{Walley91}. It produces a set
$\mathcal{P}(\varepsilon,P)$ of probabilities $P^{\ast}=(p_{1}^{\ast
},...,p_{C}^{\ast})$ such that $p_{i}^{\ast}=(1-\varepsilon)p_{i}+\varepsilon
h_{i}$, where $P=(p_{1},...,p_{C})$ is an elicited probability distribution
(in the considered case of SHAP, these distributions are $P_{S}$, $P_{S,i}$,
$P_{N}$); $h_{i}\geq0$ is arbitrary with $h_{1}+...+h_{C}=1$; $0\leq
\varepsilon\leq1$. Parameter $\varepsilon$ controls the size of set
$\mathcal{P}(\varepsilon,P)$ and can be defined from size $n$ of the training
set. The greater the number of training examples, the less the uncertainty for
probability distribution and the less the value of parameter $\varepsilon$.
The set $\mathcal{P}(\varepsilon,P)$ is a subset of the unit simplex $S(1,C)$.
Moreover, it coincides with the unit simplex when $\varepsilon=1$. According
to the model, $\mathcal{P}(\varepsilon,P)$ is the convex set of probabilities
with lower bound $(1-\varepsilon)p_{i}$ and upper bound $(1-\varepsilon
)p_{i}+\varepsilon$, i.e.,
\begin{equation}
(1-\varepsilon)p_{i}\leq p_{i}^{\ast}\leq(1-\varepsilon)p_{i}+\varepsilon
,\ i=1,...,C.
\end{equation}
The convex set has $C$ extreme points, which are all of the same form: the
$k$-th element is given by $(1-\varepsilon)p_{i}+\varepsilon$ and the other
$T-1$ elements are equal to $(1-\varepsilon)p_{i}$, i.e.,%
\begin{equation}
p_{k}^{\ast}=(1-\varepsilon)p_{k}+\varepsilon,\ p_{i}^{\ast}=(1-\varepsilon
)p_{i},\ i=1,...,C,\ i\neq k.\label{SHAP_prob_41}%
\end{equation}

\begin{figure}
[ptb]
\begin{center}
\includegraphics[
height=2.0738in,
width=2.7406in
]%
{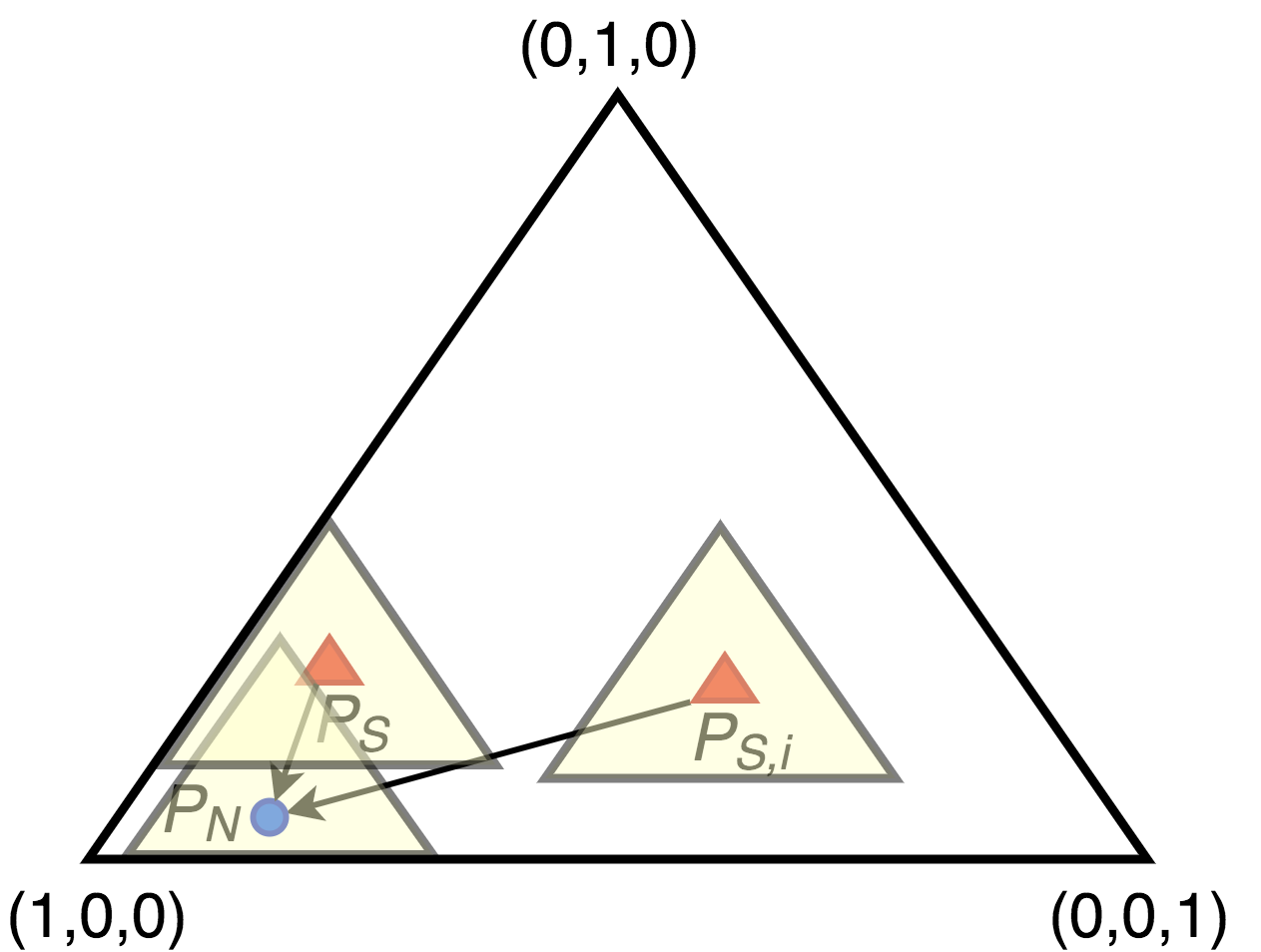}%
\caption{Illustration of distances between subsets produced by the imprecise
$\varepsilon$-contamination model}%
\label{fig:SHAP_simplex_4}%
\end{center}
\end{figure}

Fig. \ref{fig:SHAP_simplex_4} illustrates sets $\mathcal{P}(\varepsilon
,P_{S})$, $\mathcal{P}(\varepsilon,P_{S,i})$, $\mathcal{P}(\varepsilon,P_{N})$
for precise probability distributions $P_{S}$, $P_{S,i}$, $P_{N}$,
respectively, in the form of large triangles around the corresponding
distributions. Every distribution belonging to the sets can be a candidate for
some \textquotedblleft true\textquotedblright\ distribution which is actually
unknown. The imprecise $\varepsilon$-contamination model is equivalent to the
imprecise Dirichlet model (IDM) \cite{Walley96a} to some extent. The
IDM\ defined by \cite{Walley96a} can be viewed as the set of all Dirichlet
distributions over $\pi=(\pi_{1},...,\pi_{C})$ with parameters $\alpha
=(\alpha_{1},...,\alpha_{C})$ and $s$ such that the vector $\alpha$ belongs to
the unit simplex and every $\alpha_{i}$ is the mean of $\pi_{i}$ under the
Dirichlet prior. Here $\pi$ is the probability distribution defined on $C$
events, for which the Dirichlet $(s,\alpha)$ prior is defined. For the IDM,
the hyperparameter $s$ determines how quickly upper and lower probabilities of
events converge as statistical data accumulate. Smaller values of $s$ produce
faster convergence and stronger conclusions, whereas large values of $s$
produce more cautious inferences. However, hyperparameter $s$ should not
depend on the number of observations. Let $A$ be any non-trivial subset of a
sample space, and let $n(A)$ denote the observed number of occurrences of $A$
in the $n$ trials. Then, according to \cite{Walley96a}, the predictive
probability $P(A,s)$ under the Dirichlet posterior distribution is in the
following interval%
\begin{equation}
\underline{P}(A,s)=\frac{n(A)}{n+s},\ \overline{P}(A,s)=\frac{n(A)+s}{n+s}.
\end{equation}

The hyperparameter $s$ of the IDM and parameter $\varepsilon$ of the imprecise
$\varepsilon$-contamination model are connected as
\begin{equation}
\varepsilon=\frac{s}{n+s}.
\end{equation}

Hence, one can see how $\varepsilon$ depends on $n$ by a fixed value of $s$.
In particular, $\varepsilon=1$, when $n=0$, and $\varepsilon\rightarrow0$,
when $n\rightarrow\infty$.%

\subsection{Distances between subsets of probability distributions}

In order to apply the SHAP method taking into account the imprecision of the
class probability distributions, it is necessary to introduce the distance
between probability distributions and to define a way for computing the
largest and the smallest distances between all pairs of probability
distributions which belong to the small subsets of distributions
$\mathcal{P}(\varepsilon,P_{S})$, $\mathcal{P}(\varepsilon,P_{S,i})$,
$\mathcal{P}(\varepsilon,P_{N})$.

There are a lot of distance measures between probability distributions. One of
the most popular in machine learning measure is the Kullback-Leibler (KL)
divergence or relative information \cite{Kullback-Leibler-1951}:
\[
KL(P||Q)=\sum_{i=1}^{C}p_{i}\log\left(  \frac{p_{i}}{q_{i}}\right)  .
\]

This measure among other interesting measures, including relative J-divergence
\cite{Dragomir-etal-2001}, relative Arithmetic-Geometric divergence
\cite{Taneja-Kumar-06}, $\chi^{2}$-divergence \cite{Pearson-1900}, etc.
\cite{Jain-Chhabra-14,Taneja-Kumar-06}, can be derived using properties of
Csiszar's $f$-divergence measure \cite{Csiszar-1967} which is of the form:
\[
C_{f}(P||Q)=\sum_{i=1}^{C}q_{i}f\left(  \frac{p_{i}}{q_{i}}\right)  ,
\]
where the function $f:(0,\infty)\rightarrow\mathbb{R}$ is convex and
normalized, i.e., $f(1)=0$, and it is assumed that $0\cdot f(0/0)=0$ and
$0\cdot f(p/0)=0=\lim_{q\rightarrow0^{+}}qf(p/q)$. For example, functions
$f(u)=-\ln u$ and $f(u)=(1-u)^{2}/u$ generate the KL-divergence and the
$\chi^{2}$-divergence \cite{Nielsen-Nock-14}, respectively.

\subsection{A direct method to compute bounds for Shapley values}

All the aforementioned distance measures lead to very complex optimization
problems for computing bounds for distances between subsets of probability
distributions. In order to avoid solving the optimization problems, Shapley
values can be computed by means of generating a lot of points in subsets
$\mathcal{P}(\varepsilon,P_{S})$, $\mathcal{P}(\varepsilon,P_{S,i})$,
$\mathcal{P}(\varepsilon,P_{N})$ and computing the smallest and the largest
distances or computing the smallest and the largest Shapley values.

There are many algorithms for generating uniformly distributed random points
in the unit simplex
\cite{Onn-Weissman-2011,Rubinstein-Kroese2008,Rubinstein-Melamed-98}. One of
the well-known and simple algorithms is of the form
\cite{Rubinstein-Melamed-98}:

\begin{enumerate}
\item Generate $k$ independent unit-exponential random variables
$Y_{1},...,Y_{k}$ and compute $T_{k}=\sum_{i=1}^{k}Y_{i}$.

\item Define $E_{i}=Y_{i}/T_{k}$ and return vector $E=(E_{1},...,E_{k})$ which
is uniformly distributed in the unit simplex.
\end{enumerate}

Note that set $\mathcal{P}$ is convex, i.e., it is generated by finitely many
linear constraints. This implies that it is totally defined by its extreme
points or vertices denoted $\mathcal{E}(\mathcal{P})$. Suppose that we have a
set of $r$ extreme points $Q_{k}=(q_{1}^{(k)},...,q_{C}^{(k)})$ of the set
$\mathcal{P}$, i.e., $q_{k}\in\mathcal{E}(\mathcal{P})$, $k=1,...,r$. In the
case of the imprecise $\varepsilon$-contamination model, there holds $r=C.$
Then every probability distribution $P=(p_{1},...,p_{C})$ from $\mathcal{P}$
can be represented as the linear combination of the extreme points
\begin{equation}
P=\sum_{k=1}^{r}\lambda_{k}\cdot Q_{k}.\label{Ada_extr_21}%
\end{equation}

Here $\lambda=(\lambda_{1},...,\lambda_{r})$ is a vector of weights such that
$\lambda_{1}+...+\lambda_{r}=1$.

If to uniformly generate vectors $\lambda$ from the unit simplex by using one
of the known algorithms of generation, then random points $P$ from the set
$\mathcal{P}$ can be obtained. Extreme points of set $\mathcal{P}$ for the
imprecise $\varepsilon$-contamination model are given in (\ref{SHAP_prob_41}).

From all generated points in $\mathcal{P}(\varepsilon,P_{S})$, $\mathcal{P}%
(\varepsilon,P_{S,i})$, $\mathcal{P}(\varepsilon,P_{N})$, we select three
corresponding points, say, $P_{1}\in\mathcal{P}(\varepsilon,P_{S})$, $R_{1}%
\in\mathcal{P}(\varepsilon,P_{S,i})$, $Q_{1}\in\mathcal{P}(\varepsilon,P_{N})$
such that the difference $D(P,Q)-D(R,Q)$ of distances achieves its minimum.
These points contribute into the lower Shapley value $\phi_{i}^{L}$. In the
same way, three points $P_{2}$, $R_{2}$, $Q_{2}$ are selected from the same
sets, respectively, such that the difference $D(P,Q)-D(R,Q)$ achieves its
maximum. These points contribute into the upper Shapley value $\phi_{i}^{U}$.
This procedure is repeated for all Shapley values $\phi_{i}$, $i=1,...,m$.

The main advantage of the above generation approach for computing the lower
and upper Shapley values is that an arbitrary distance between probability
distributions can be used for implementing the approach, for example, the KL
divergence or the $\chi^{2}$-divergence. There are no restrictions for
selecting distances. Moreover, intervals for Shapley values can be also
directly computed. Of course, this approach may be extremely complex
especially when training examples have a high dimension and parameter
$\varepsilon$ of the imprecise $\varepsilon$-contamination model is rather
large (the imprecise model produces large sets $\mathcal{P}(\varepsilon
,\cdot)$). Moreover, it is difficult to control the accuracy of the obtained
Shapley values because there is a chance that the selected points $P_{1}$,
$R_{1}$, $Q_{1}$ and $P_{2}$, $R_{2}$, $Q_{2}$ may be not optimal, i.e., the
corresponding differences $D(P,Q)-D(R,Q)$ do not achieve the minimum and the
maximum. In order to overcome this difficulty, another approach is proposed,
which is based on applying the Kolmogorov-Smirnov distance leading to a set of
very simple linear programming problems.

\section{The Kolmogorov-Smirnov distance and the imprecise SHAP}

The Kolmogorov-Smirnov distance between two probability distributions
$P=(p_{1},...,p_{C})$ and $Q=(q_{1},...,q_{C})$ is defined as the maximal
distance between the cumulative distributions. Let $\pi=(\pi_{1},...,\pi_{C})$
and $\alpha=(\alpha_{1},...,\alpha_{C})$ be cumulative probability
distributions corresponding to the distributions $P$ and $Q$, respectively,
where $\pi_{i}=\sum_{j=1}^{i}p_{j}$ and $\alpha_{i}=\sum_{j=1}^{i}q_{j}$. It
is assumed that $\pi_{C}=\alpha_{C}=1$. Then the Kolmogorov-Smirnov distance
is of the form:
\begin{equation}
D_{KS}(P,Q)=\max_{i=1,..,C-1}\left\vert \pi_{i}-\alpha_{i}\right\vert .
\label{SHAP_prob_51}%
\end{equation}

An imprecise extension of the Kolmogorov-Smirnov distance with using the
imprecise $\varepsilon$-contamination model has been studied by Montes et al.
\cite{Montes-etal-20}. However, we consider the corresponding distances as
elements of optimization problems for computing lower and upper bounds for
Shapley values.

First, we extend the definition of the imprecise $\varepsilon$-contamination
model $p_{i}^{\ast}=(1-\varepsilon)p_{i}+\varepsilon h_{i}$ on the case of
probabilities of arbitrary events $A$. The model defines $P^{\ast}(A)$ in the
same way, i.e.,
\begin{equation}
P^{\ast}(A)=(1-\varepsilon)P(A)+\varepsilon H(A),
\end{equation}
where $P(A)$ is an elicited probability of event $A$, $H(A)$ is the
probability of $A$ under condition of arbitrary distribution.

This implies that bounds for $P^{\ast}(A)$ are defined as follows:
\begin{equation}
(1-\varepsilon)P(A)\leq P^{\ast}(A)\leq(1-\varepsilon)P(A)+\varepsilon.
\label{SHAP_prob_52}%
\end{equation}

If $A=\{1,2,...,i\}$, then $P^{\ast}(A)$ is the $i$-th cumulative probability.
Hence, we can define bounds for cumulative probability distributions. Denote
sets of cumulative probability distributions $\pi$, $\tau$, $\alpha$ produced
by the imprecise $\varepsilon$-contamination model for the distributions
$P_{S}$, $P_{S,i}$, $P_{N}$ as $\mathcal{R}(\varepsilon,P_{S})$,
$\mathcal{R}(\varepsilon,P_{S,i})$, $\mathcal{R}(\varepsilon,P_{N})$,
respectively. It is supposed that the lower and upper bounds for probabilities
$\pi_{i}$, $\tau_{i}$, $\alpha_{i}$, $i=1,...,C-1$, are
\begin{equation}
\pi_{i}^{L}\leq\pi_{i}\leq\pi_{i}^{U},\ \ \tau_{i}^{L}\leq\tau_{i}\leq\tau
_{i}^{U},\ \ \alpha_{i}^{L}\leq\alpha_{i}\leq\alpha_{i}^{U}%
.\label{SHAP_prob_54}%
\end{equation}

We assume that $\pi_{C}=\tau_{C}=\alpha_{C}=1$ because these values represent
the cumulative distribution function.

Let us consider the lower bound $L$ for $D_{KS}(P,Q)-D_{KS}(R,Q)$. It can be
derived from the following optimization problem:
\begin{equation}
L=\min_{\pi,\tau,\alpha}\left(  \max_{i=1,..,C-1}\left\vert \pi_{i}-\alpha
_{i}\right\vert -\max_{i=1,..,C-1}\left\vert \tau_{i}-\alpha_{i}\right\vert
\right)  , \label{SHAP_prob_55}%
\end{equation}
subject to (\ref{SHAP_prob_54}).

This is non-convex optimization problem. However, it can be represented as a
set of $2(C-1)$ simple linear programming problems.

\begin{proposition}
\label{prop:Low_Dist}The lower bound $L$ for $D_{KS}(P,Q)-D_{KS}(R,Q)$ or the
solution of problem (\ref{SHAP_prob_55}) is determined by solving $2(C-1)$
linear programming problems:
\begin{equation}
L_{1}(k)=\min_{B,\pi,\alpha}\left(  B-\tau_{k}^{U}+\alpha_{k}\right)
,\ k=1,...,C-1, \label{SHAP_prob_60}%
\end{equation}%
\begin{equation}
L_{2}(k)=\min_{B,\pi,\alpha}\left(  B-\alpha_{k}+\tau_{k}^{L}\right)
,\ k=1,...,C-1, \label{SHAP_prob_61}%
\end{equation}
subject to
\begin{equation}
\pi_{i}^{L}\leq\pi_{i}\leq\pi_{i}^{U},\alpha_{i}^{L}\leq\alpha_{i}\leq
\alpha_{i}^{U},\ i=1,...,C-1, \label{SHAP_prob_56}%
\end{equation}%
\begin{equation}
B\geq\pi_{i}-\alpha_{i},\ B\geq\alpha_{i}-\pi_{i},\ i=1,...,C-1,
\label{SHAP_prob_57}%
\end{equation}%
\begin{equation}
\pi_{i}\leq\pi_{i+1},\ \alpha_{i}\leq\alpha_{i+1},\ i=1,...,C-2,
\label{SHAP_prob_58}%
\end{equation}
For $L_{1}(k)$, the following constraints are added:%
\begin{equation}
\alpha_{k}\leq\tau_{k}^{U}~\text{and }\tau_{k}^{U}-\alpha_{k}\geq\tau_{i}%
^{U}-\alpha_{i},i=1,...,C-1,\ i\neq k. \label{SHAP_prob_59}%
\end{equation}
For $L_{2}(k)$, then the following constraints are added:
\begin{equation}
\alpha_{k}\geq\tau_{k}^{L}~\text{and }\alpha_{k}-\tau_{k}^{L}\geq\alpha
_{i}-\tau_{i}^{L},i=1,...,C-1,\ i\neq k. \label{SHAP_prob_59_2}%
\end{equation}

A final solution is determined by selecting $k_{1}=\arg\max_{k}\left(
\tau_{k}^{U}-\alpha_{k}\right)  $ from (\ref{SHAP_prob_60}) and $k_{2}%
=\arg\max_{k}\left(  \alpha_{k}-\tau_{k}^{L}\right)  $ from
(\ref{SHAP_prob_61}). The lower bound is
\begin{equation}
L=\left\{
\begin{array}
[c]{cc}%
L_{1}(k_{1}), & \text{if }\tau_{k_{1}}^{U}-\alpha_{k_{1}}\geq\alpha_{k_{2}%
}-\tau_{k_{2}}^{L},\\
L_{2}(k_{2}), & \text{otherwise.}%
\end{array}
\right.  \label{SHAP_prob_55_1}%
\end{equation}

\end{proposition}

It should be noted that problems (\ref{SHAP_prob_60}) do not have solutions
for some $k$ when inequality $\tau_{k}^{U}<\alpha_{k}^{L}$ is valid. This
follows from constraints $\alpha_{k}\leq\tau_{k}^{U}$ in (\ref{SHAP_prob_59})
and $\alpha_{k}^{L}\leq\alpha_{k}$ in (\ref{SHAP_prob_56}) . The same can be
said about solutions of problems (\ref{SHAP_prob_61}). They do not have
solutions when inequality $\tau_{k}^{L}>\alpha_{k}^{U}$ is valid. This follows
from constraints $\alpha_{k}\geq\tau_{k}^{L}$ in (\ref{SHAP_prob_59_2}) and
$\alpha_{k}\leq\alpha_{k}^{U}$ in (\ref{SHAP_prob_56}).

Let us consider now the upper bound $U$ for $D_{KS}(P,Q)-D_{KS}(R,Q)$. It can
be derived from the following optimization problem:
\begin{equation}
U=\max_{\pi,\tau,\alpha}\left(  \max_{i=1,..,C-1}\left\vert \pi_{i}-\alpha
_{i}\right\vert -\max_{i=1,..,C-1}\left\vert \tau_{i}-\alpha_{i}\right\vert
\right)  , \label{SHAP_prob_65}%
\end{equation}
subject to (\ref{SHAP_prob_54}).

\begin{proposition}
\label{prop:Upp_Dist}The upper bound $U$ for $D_{KS}(P,Q)-D_{KS}(R,Q)$ or the
solution of problem (\ref{SHAP_prob_65}) is determined by solving $2C-1$
linear programming problems:
\begin{equation}
U_{1}(k)=\max_{B,\tau,\alpha}\left(  \pi_{k}^{U}-\alpha_{k}-B\right)
,\ k=1,...,C-1, \label{SHAP_prob_70}%
\end{equation}%
\begin{equation}
U_{2}(k)=\max_{B,\tau,\alpha}\left(  \alpha_{k}-\pi_{k}^{L}-B\right)
,\ k=1,...,C-1, \label{SHAP_prob_71}%
\end{equation}
subject to
\begin{equation}
\tau_{i}^{L}\leq\tau_{i}\leq\tau_{i}^{U},\alpha_{i}^{L}\leq\alpha_{i}%
\leq\alpha_{i}^{U},\ i=1,...,C-1, \label{SHAP_prob_76}%
\end{equation}%
\begin{equation}
B\geq\tau_{i}-\alpha_{i},\ B\geq\alpha_{i}-\tau_{i},\ i=1,...,C-1,
\end{equation}%
\begin{equation}
\tau_{i}\leq\tau_{i+1},\ \alpha_{i}\leq\alpha_{i+1},\ i=1,...,C-2,
\end{equation}
For $L_{1}(k)$, the following constraints are added:%
\begin{equation}
\alpha_{k}\leq\pi_{k}^{U}~\text{and }\pi_{k}^{U}-\alpha_{k}\geq\pi_{i}%
^{U}-\alpha_{i},~i=1,...,C-1,\ i\neq k. \label{SHAP_prob_79}%
\end{equation}
For $L_{2}(k)$, the following constraints are added:
\begin{equation}
\alpha_{k}\geq\pi_{k}^{L}~\text{and }\alpha_{k}-\pi_{k}^{L}\geq\alpha_{i}%
-\pi_{i}^{L},~i=1,...,C-1,\ i\neq k. \label{SHAP_prob_79_2}%
\end{equation}

A final solution is determined by selecting $k_{1}=\arg\max_{k}\left(  \pi
_{k}^{U}-\alpha_{k}\right)  $ from (\ref{SHAP_prob_70}) and $k_{2}=\arg
\max_{k}\left(  \alpha_{k}-\pi_{k}^{L}\right)  $ from (\ref{SHAP_prob_71}).
The upper bound is
\begin{equation}
U=\left\{
\begin{array}
[c]{cc}%
U_{1}(k_{1}), & \text{if }\pi_{k_{1}}^{U}-\alpha_{k_{1}}\geq\alpha_{k_{2}}%
-\pi_{k_{2}}^{L},\\
U_{2}(k_{2}), & \text{otherwise.}%
\end{array}
\right.
\end{equation}

\end{proposition}

It should be noted that problems (\ref{SHAP_prob_70}) do not have solutions
for some $k$ when inequality $\pi_{k}^{U}<\alpha_{k}^{L}$ is valid. This
follows from constraints $\alpha_{k}^{L}\leq\alpha_{k}$ in (\ref{SHAP_prob_76}%
) and $\alpha_{k}\leq\pi_{k}^{U}$ in (\ref{SHAP_prob_79}). The same can be
said about solutions of problems (\ref{SHAP_prob_71}). They do not have
solutions when inequality $\pi_{k}^{L}\geq\alpha_{k}^{U}$ is valid. This
follows from constraints $\alpha_{k}\leq\alpha_{k}^{U}$ in (\ref{SHAP_prob_76}%
) and $\alpha_{k}\geq\pi_{k}^{L}$ in (\ref{SHAP_prob_79_2}).

The Kolmogorov-Smirnov distance for the binary classification black-box model
is of the form:%
\begin{equation}
D_{KS}(P,Q)=\left\vert \pi_{1}-\alpha_{1}\right\vert .
\end{equation}

\begin{equation}
L=\min_{\pi,\tau,\alpha}\left(  \left\vert \pi_{1}-\alpha_{1}\right\vert
-\left\vert \tau_{1}-\alpha_{1}\right\vert \right)  ,
\end{equation}

The index $1$ will be omitted below for short.

\begin{corollary}
\label{cor:SHAP1}For the binary classification black-box model with $C=2$, the
lower $L$ and upper $U$ bounds for $D_{KS}(P,Q)-D_{KS}(R,Q)$ are
\begin{equation}
L=\left\{
\begin{array}
[c]{cc}%
2\alpha^{L}-\pi^{U}-\tau^{U}, & \text{if }\pi^{U}\leq\alpha^{L}\leq\tau^{U},\\
\pi^{L}-\tau^{U}, & \text{if }\alpha^{L}\leq\min(\tau^{U},\pi^{L}),\\
\tau^{L}-\pi^{U}, & \text{if }\alpha^{U}\geq\max(\tau^{L},\pi^{U}),\\
\pi^{L}+\tau^{L}-2\alpha^{U}, & \text{if }\tau^{L}\leq\alpha^{U}\leq\pi^{L},
\end{array}
\right.  \label{SHAP_prob_83}%
\end{equation}%
\begin{equation}
U=\left\{
\begin{array}
[c]{cc}%
2\alpha^{U}-\pi^{L}-\tau^{L}, & \text{if }\pi^{L}\leq\alpha^{U}\leq\tau^{L},\\
\pi^{U}-\tau^{L}, & \text{if }\alpha^{U}\leq\min(\tau^{L},\pi^{U}),\\
\tau^{U}-\pi^{L}, & \text{if }\alpha^{L}\geq\max(\tau^{U},\pi^{L}),\\
\pi^{U}+\tau^{U}-2\alpha^{L}, & \text{if }\tau^{U}\leq\alpha^{L}\leq\pi^{U}.
\end{array}
\right.  . \label{SHAP_prob_84}%
\end{equation}

\end{corollary}

It follows from (\ref{SHAP_prob_83}) and (\ref{SHAP_prob_84}) that there holds
for the case of precise probabilities:%
\begin{equation}
D_{KS}(P,Q)-D_{KS}(R,Q)=\left\vert \pi-\alpha\right\vert -\left\vert
\tau-\alpha\right\vert .
\end{equation}

The next question is how to compute bounds $D^{L}$ and $D^{U}$ for $D$ in
(\ref{SHAP_prob_25}), which are used to reduce intervals of Shapley values.

Let $\pi$ and $\alpha$ be cumulative distribution functions corresponding to
$P_{\emptyset}$ and $P_{N}$ defined in (\ref{SHAP_prob_11}), respectively,
i.e., $\pi\in P_{\emptyset}$ and $\alpha\in P_{N}$. Then the lower bound
$D^{L}$ is determined
\begin{equation}
D^{L}=\min_{\pi,\alpha}\max_{i=1,...,C-1}\left\vert \pi_{i}-\alpha
_{i}\right\vert .
\end{equation}

The upper bound $D^{U}$ is determined from the following optimization problem:%
\begin{equation}
D^{U}=\max_{\pi,\alpha}\max_{i=1,...,C-1}\left\vert \pi_{i}-\alpha
_{i}\right\vert .
\end{equation}

\begin{proposition}
\label{prop:general_bounds}The lower bound $D^{L}$ is determined from the
following linear programming problem:
\begin{equation}
D^{L}=\min_{\pi,\alpha}B,
\end{equation}
subject to (\ref{SHAP_prob_56})-(\ref{SHAP_prob_58}).

The upper bound $D^{U}$ is determined as
\begin{equation}
D^{U}=\max_{k=1,...,C-1}\max\left(  \pi_{k}^{U}-\alpha_{k}^{L},\alpha_{k}%
^{U}-\pi_{k}^{L}\right)  .
\end{equation}

\end{proposition}

Finally, we have obtained basic tools for dealing with imprecise Shapley
values, which include Propositions \ref{prop:Low_Dist}, \ref{prop:Upp_Dist},
and \ref{prop:general_bounds}.

\section{Numerical experiments}

In all numerical experiments, the black-box model is the random forest
consisting of $100$ decision trees with largest depth $8$. The choice of the
random forest is due to simplicity of getting the class probabilities.

\subsection{Numerical experiments with synthetic data}

In order to study properties of the imprecise SHAP and to compare its
predictions with the original SHAP, we generate three datasets of synthetic
data. Every dataset consists of $1000$ training examples and $250$ testing
examples such that every example is characterized by two features ($x$ and
$y$). The datasets are described in Table \ref{t:imp_SHAP_datasets_1} where
the second column shows the number of classes, numbers of training and testing
examples in the classes are given in the third and fourth columns, respectively.%

\begin{table}[tbp] \centering
\caption{A brief introduction about datasets}%
\begin{tabular}
[c]{cccc}\hline
Dataset & Classes & Training & Testing\\\hline
1 & 2 & $879,121$ & $215,35$\\\hline
2 & 3 & $340,336,324$ & $92,80,78$\\\hline
3 & 4 & $249,263,238,250$ & $64,50,74,62$\\\hline
\end{tabular}
\label{t:imp_SHAP_datasets_1}%
\end{table}%

\begin{figure}
[ptb]
\begin{center}
\includegraphics[
height=2.194in,
width=5.5434in
]%
{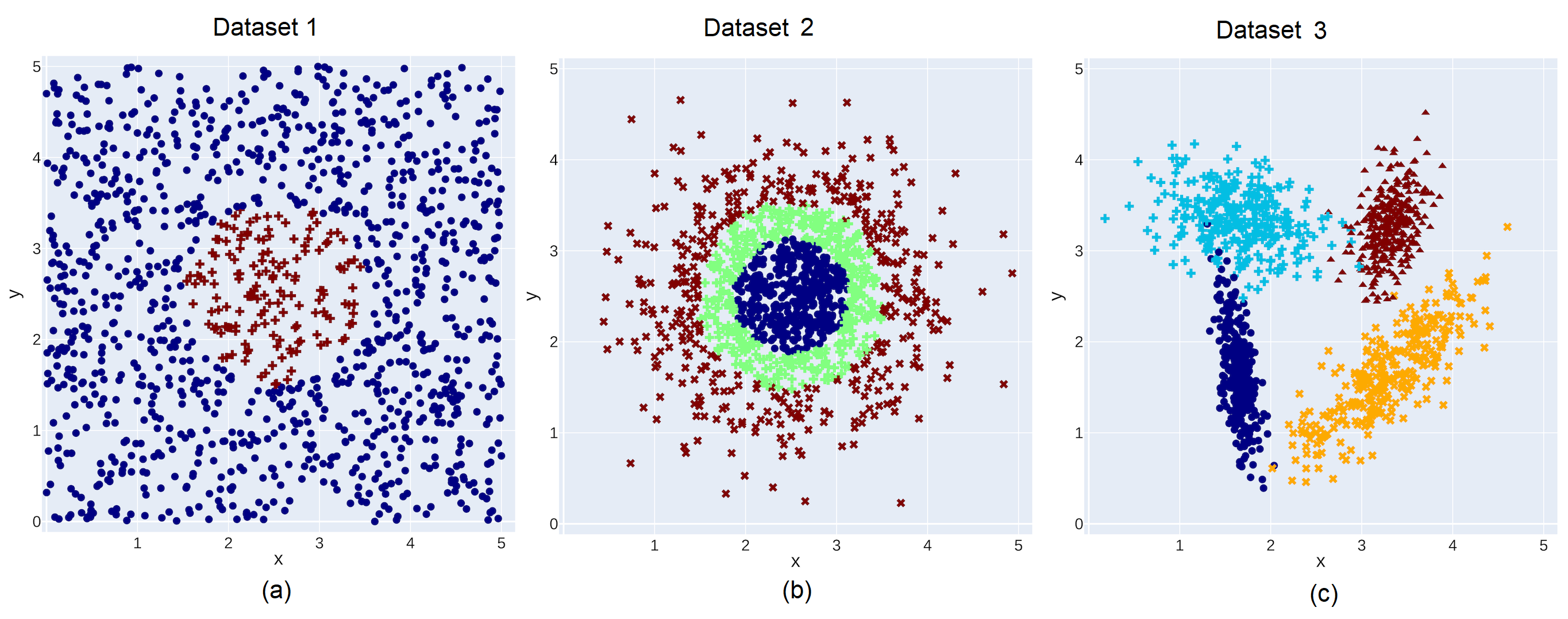}%
\caption{Three synthetic datasets for numerical experiments}%
\label{f:synt_data_join}%
\end{center}
\end{figure}

Every example of the first dataset is randomly generated from the uniform
distribution in $[0,5]\times\lbrack0,5]$. Points inside a circle with the unit
radius and center $(2.5,2.5)$ belongs to the first class, other points belongs
to the second class. Points of the dataset are shown in Fig.
\ref{f:synt_data_join} (a).%

Every example of the second dataset is randomly generated from the normal
distribution with the expectation $(2.5,2.5)$ and the covariance matrix
$0.5\cdot\mathbf{I}$, where $\mathbf{I}$ is the unit matrix. Points inside a
circle with the unit radius and center $(2.5,2.5)$ belongs to the first class.
Points located outside the circle with radius $2$ and center $(2.5,2.5)$
belong to the third class. Other points belong to the second class. Points of
the dataset are shown in Fig. \ref{f:synt_data_join} (b).

Every example of the third dataset is randomly generated in accordance with a
procedure implemented in the Python library \textquotedblleft
sklearn\textquotedblright\ in function \textquotedblleft
sklearn.datasets.make\_classification\textquotedblright. According to the
procedure, clusters of points normally distributed with the variance $1$ about
vertices of the square with sides of length $2$ are generated. Details of the
generating procedure can be found in
https://scikit-learn.org/stable/modules/generated/sklearn.datasets.make\_classification.html.
Points of the dataset are shown in Fig. \ref{f:synt_data_join} (c).%

\begin{figure}
[ptb]
\begin{center}
\includegraphics[
height=4.9104in,
width=3.3356in
]%
{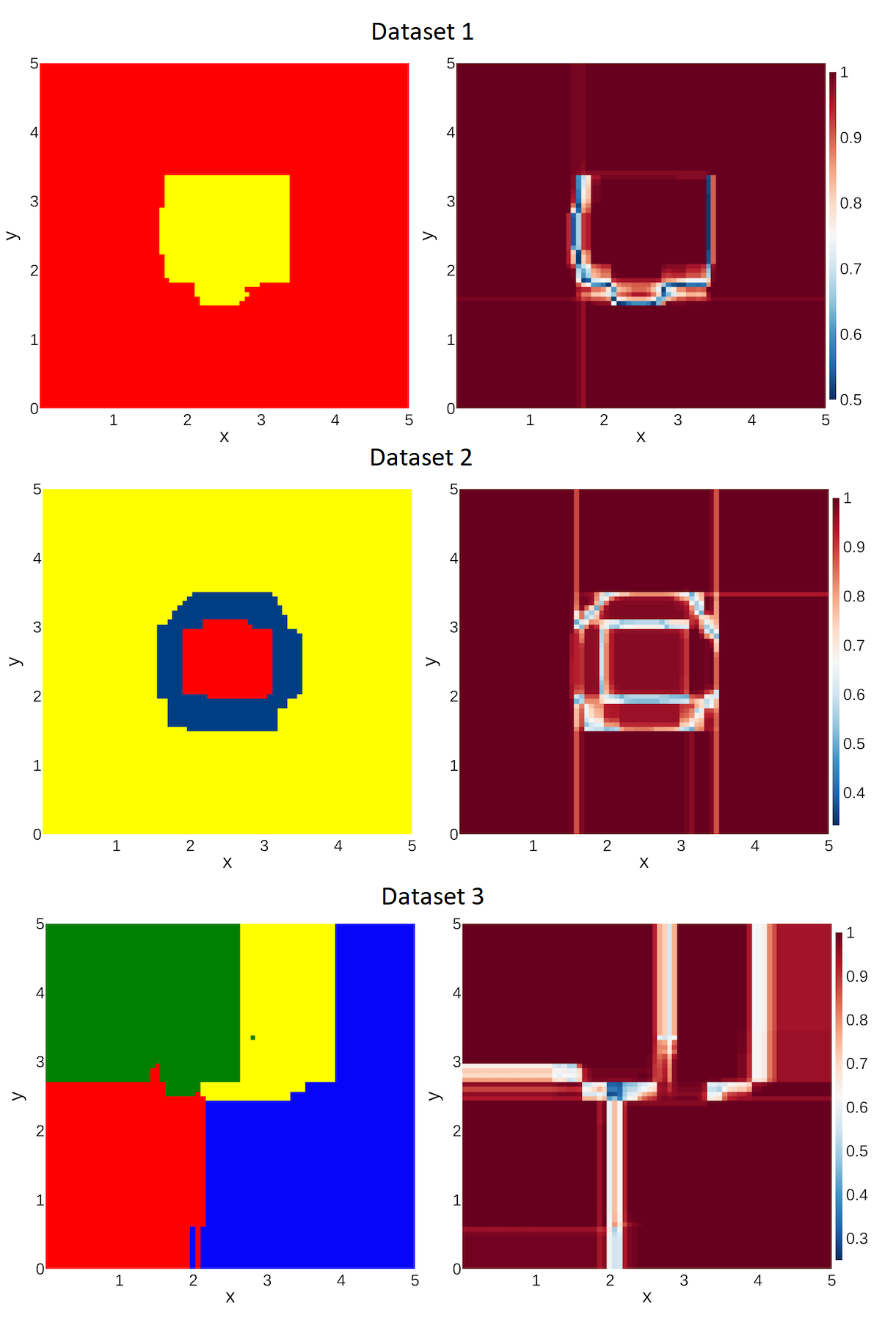}%
\caption{Predictions of the black-box random forest for three datasets (three
rows of pictures), the left picture in every row depicts predicted classes,
the right picture depicts heatmaps of largest probabilities of predicted
classes }%
\label{f:synt_data_pred}%
\end{center}
\end{figure}

Predictions of the black-box random forest for every dataset are illustrated
in Fig. \ref{f:synt_data_pred} where every row of pictures corresponds to a
certain dataset, the first column of pictures depicts predicted classes by
different colors for every dataset, the second column depicts heatmaps
corresponding to largest probabilities of predicted classes. It is clearly
seen from the second column of pictures in Fig. \ref{f:synt_data_pred} that
the black-box random forest cannot cope with examples which are close to
boundaries between classes.%

\begin{figure}
[ptb]
\begin{center}
\includegraphics[
height=2.4975in,
width=4.8329in
]%
{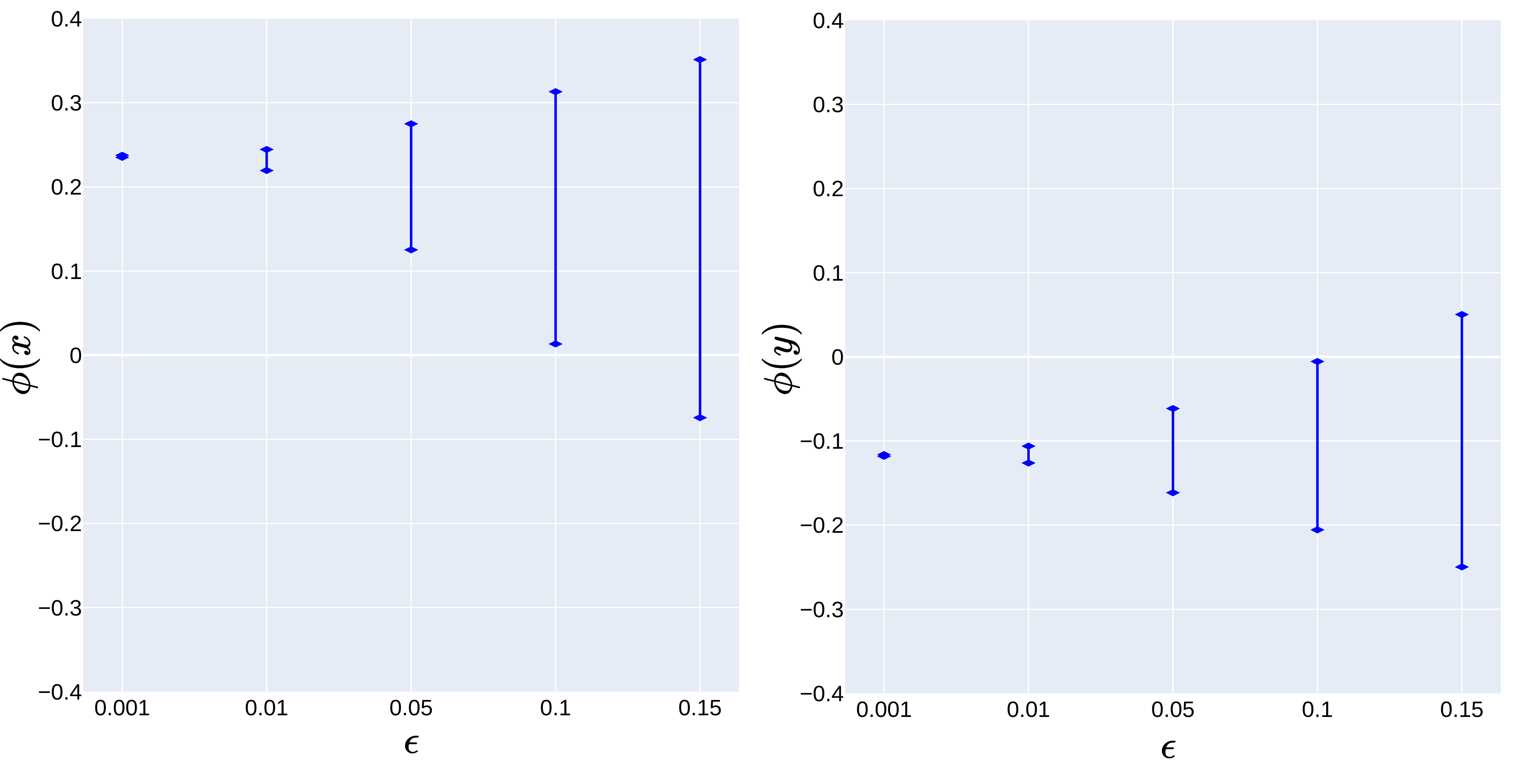}%
\caption{Intervals of Shapley values for point $(1.5,2.5)$ from the first
synthetic dataset with two features, $x$ and $y$, for different contamination
parameters $\varepsilon$}%
\label{f:point_15_25_dat_1}%
\end{center}
\end{figure}

\begin{figure}
[ptb]
\begin{center}
\includegraphics[
height=2.3131in,
width=5.0156in
]%
{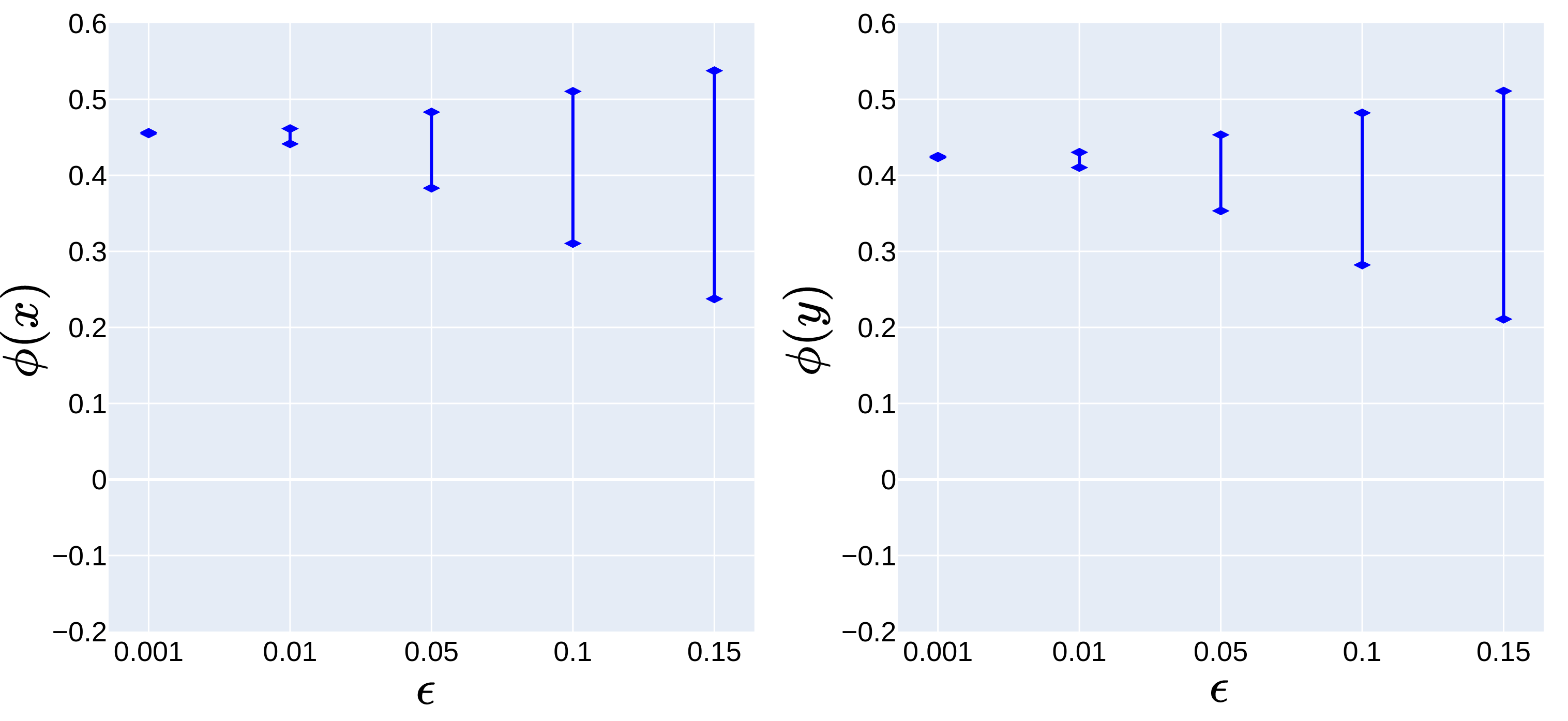}%
\caption{Intervals of Shapley values for point $(2.5,2.5)$ from the first
synthetic dataset with two features, $x$ and $y$, for different contamination
parameters $\varepsilon$}%
\label{f:both_25_25_simple}%
\end{center}
\end{figure}

An example of interval-valued Shapley values as functions of the contamination
parameter $\varepsilon$ is shown in Fig. \ref{f:point_15_25_dat_1}. The
testing example with features $(1.5,2.5)$ is chosen because this point is very
close to the boundary between classes and the corresponding class distribution
is non-trivial. Moreover, the location of the point simply shows that feature
$x$ is important because its small perturbation changes the class (see Fig.
\ref{f:synt_data_join} (a)) in contrast to feature $y$ which does not change
the class. One can see from Fig. \ref{f:point_15_25_dat_1} that intervals of
Shapley values for feature $x$ are clearly larger than the intervals for $y$.
At the same time, intervals for Shapley values of $x$ and $y$ are intersecting
when $\varepsilon=0.15$. However, the intersecting area is rather small,
therefore, we can confidently assert that feature $x$ is more important.
Another analyzed point is $(2.5,2.5).$ It is seen from Fig.
\ref{f:synt_data_join} (a) that this point is in the center of the circle, and
both the features equally contribute into the prediction. The same can be
concluded from Fig. \ref{f:both_25_25_simple} where intervals of Shapley
values are shown. One can see from Fig. \ref{f:both_25_25_simple} that the
obtained intervals are more narrow than the same intervals given in Fig.
\ref{f:point_15_25_dat_1}.

Moreover, the above experiments clearly show that the approach for computing
contributions of every feature based on the difference of distances
$D(P_{S},P_{N})-D(P_{S,i},P_{N})$ provides correct results which are
consistent with the visual relationship between features and predictions
depicted in Fig. \ref{f:synt_data_pred}.%

\begin{figure}
[ptb]
\begin{center}
\includegraphics[
height=2.3989in,
width=5.1031in
]%
{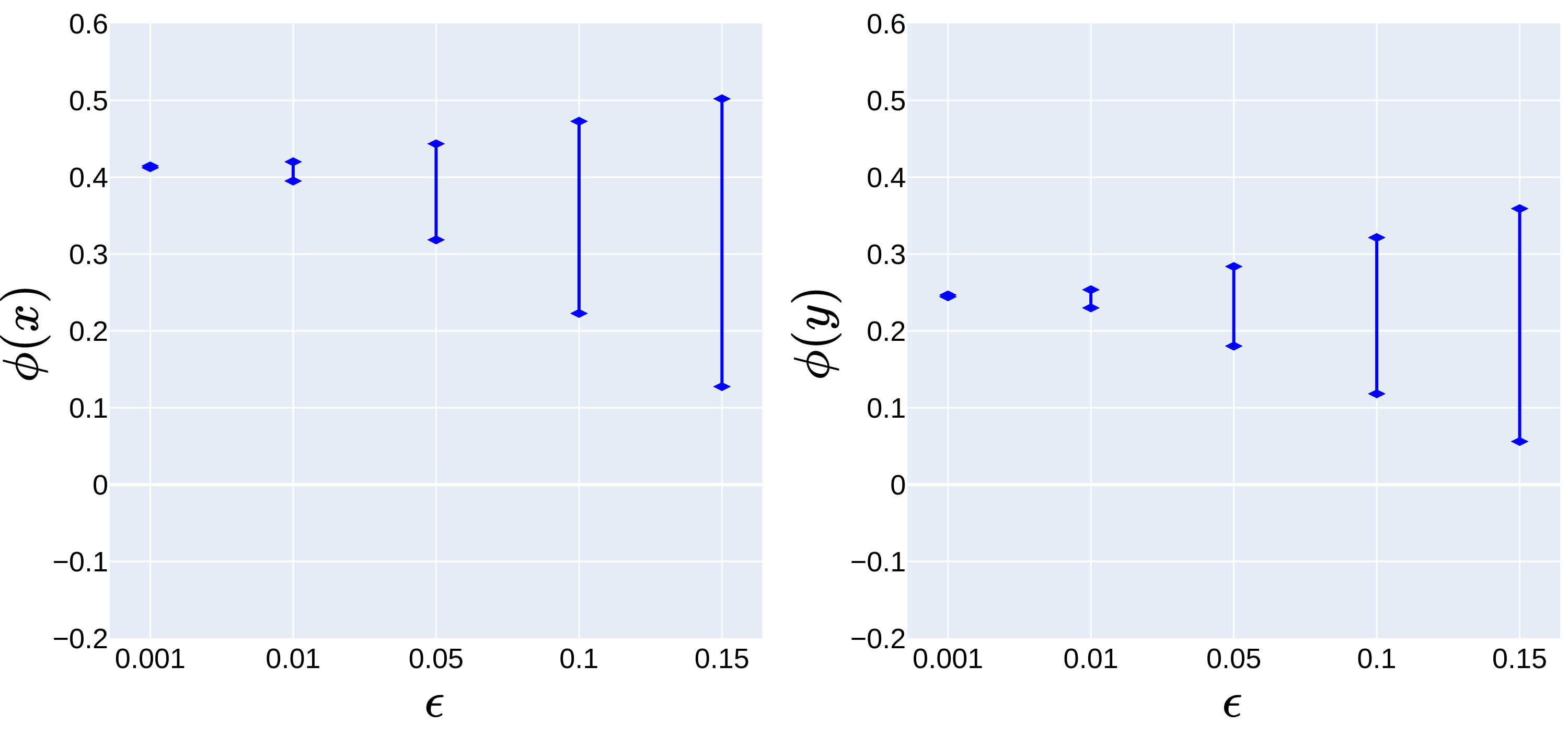}%
\caption{Intervals of Shapley values for point $(3.5,1.5)$ from the second
synthetic dataset with two features, $x$ and $y$, for different contamination
parameters $\varepsilon$}%
\label{f:both_35_15_2}%
\end{center}
\end{figure}

\begin{figure}
[ptb]
\begin{center}
\includegraphics[
height=2.4478in,
width=5.1117in
]%
{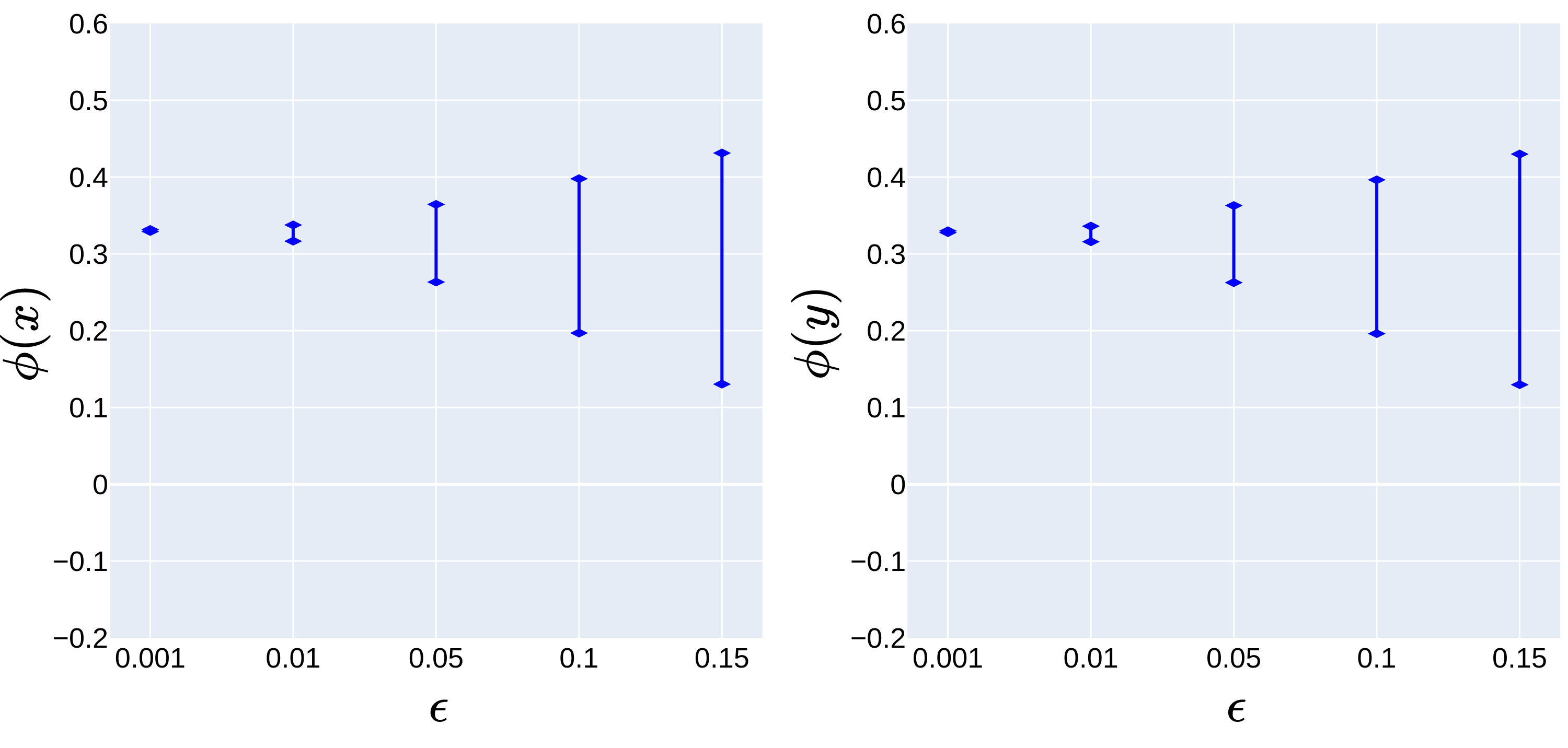}%
\caption{Intervals of Shapley values for point $(4.5,4.5)$ from the second
synthetic dataset with two features, $x$ and $y$, for different contamination
parameters $\varepsilon$}%
\label{f:both_45_45_2}%
\end{center}
\end{figure}

Similar examples can be given for the second and the third synthetic datasets.
In particular, Fig. \ref{f:both_35_15_2} shows intervals of Shapley values
corresponding two features, $x$ and $y$, when the black-box random forest
predicts point $(3.5,1.5)$ located close to the class boundary (see Fig.
\ref{f:synt_data_join} (b)). One can see again that the interval relationship
corresponds to the visual relationship between features and predictions
depicted in Fig. \ref{f:synt_data_pred}. Fig. \ref{f:both_45_45_2} shows
intervals of Shapley values for point $(4.5,4.5)$ located far from the class
boundary (see Fig. \ref{f:synt_data_join} (b)).%

%

\begin{figure}
[ptb]
\begin{center}
\includegraphics[
height=2.2917in,
width=5.074in
]%
{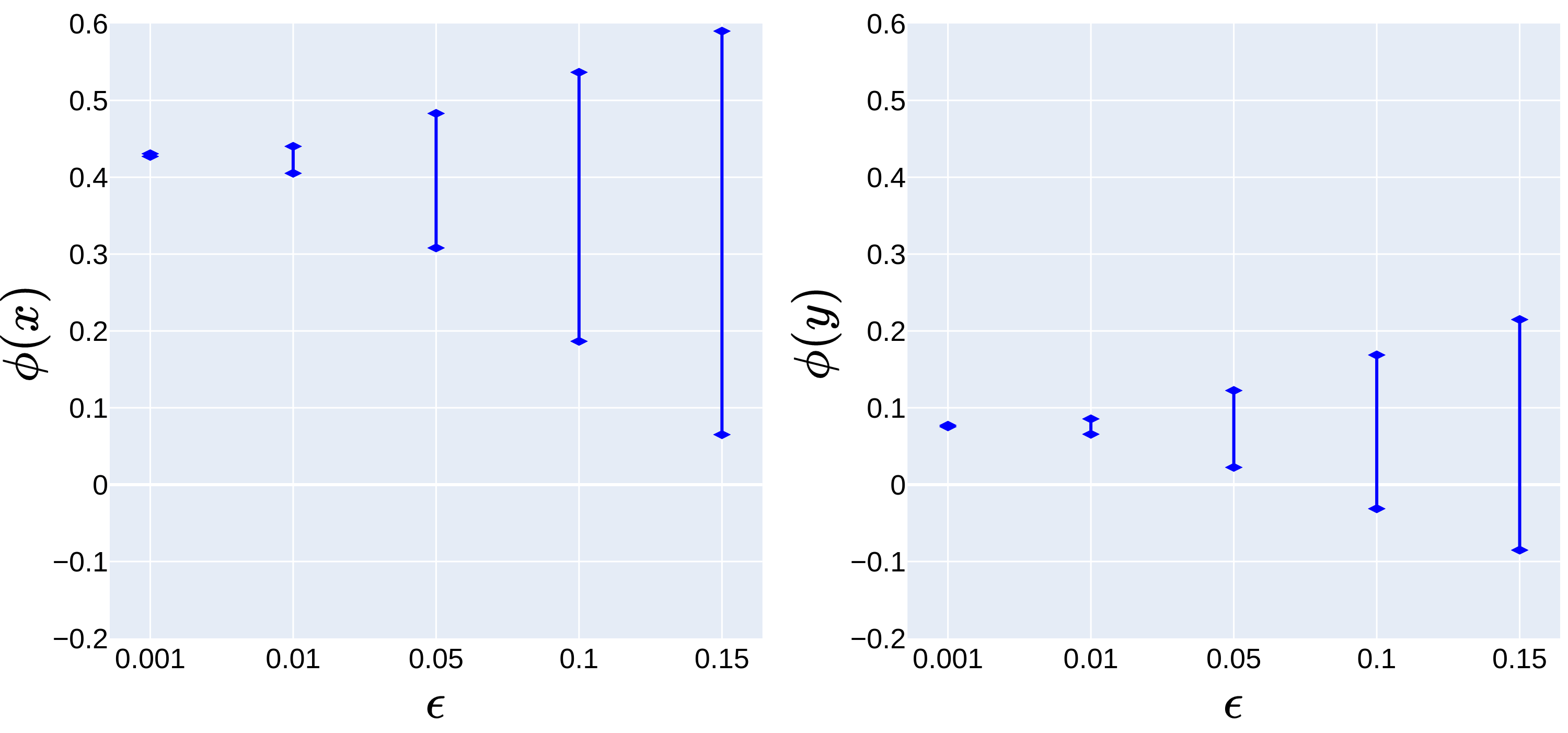}%
\caption{Intervals of Shapley values for point $(4.0,4.0)$ from the third
synthetic dataset with two features, $x$ and $y$, for different contamination
parameters $\varepsilon$}%
\label{f:both_4_4_3}%
\end{center}
\end{figure}

\begin{figure}
[ptb]
\begin{center}
\includegraphics[
height=2.3774in,
width=5.054in
]%
{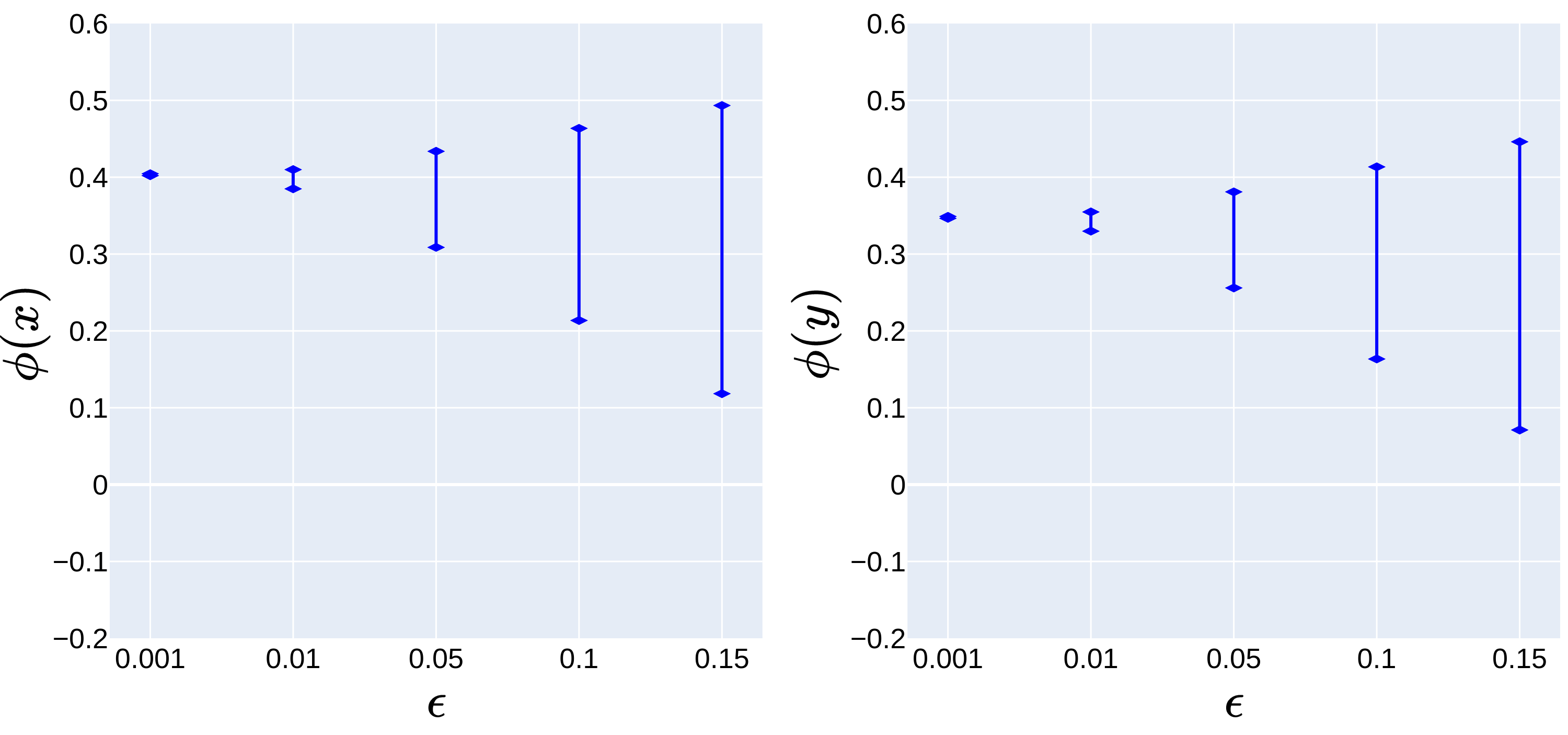}%
\caption{Intervals of Shapley values for point $(1.0,1.0)$ from the third
synthetic dataset with two features, $x$ and $y$, for different contamination
parameters $\varepsilon$}%
\label{f:both_1_1_3}%
\end{center}
\end{figure}

Figs. \ref{f:both_4_4_3}-\ref{f:both_1_1_3} show similar pictures for the
third synthetic dataset where predictions corresponding to points $(4.0,4.0)$
and $(1.0,1.0)$ are explained. It is interesting to point out that the case of
point $(4.0,4.0)$. If we look at Fig. \ref{f:synt_data_join} (c), then we see
that point belong to the area of one of the classes. This implies that the
corresponding prediction almost equally depends on $x$ and $y$. However, if we
look at Fig. \ref{f:synt_data_pred} (the third row), then we see that the
classifier does not provide a certain class as a prediction, and the explainer
selects feature $x$ as the most important one assuming that $y$ does not
change the class though this is not obvious from Fig. \ref{f:synt_data_join}
(c). This observation is justified by the fact that intervals of Shapley
values for $x$ are the largest ones for all $\varepsilon>0$, i.e., imprecision
of the corresponding prediction impacts on imprecision of Shapley values.%

%

\begin{figure}
[ptb]
\begin{center}
\includegraphics[
height=2.3791in,
width=1.9735in
]%
{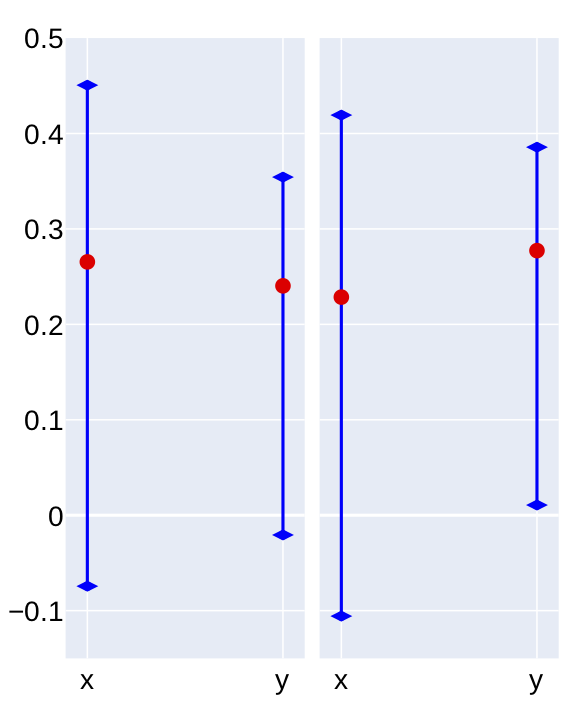}%
\caption{Intervals of Shapley values for two points $(2.8,1.5)$ and
$(1.84,3.34)$}%
\label{f:12}%
\end{center}
\end{figure}

Fig. \ref{f:12} illustrates cases of the Shapley value interval relationship
for $\varepsilon=0.15$ (blue segments) and for $\varepsilon=0$ (red circles)
for the third dataset. If to consider the precise Shapley values, then the
first case (the left picture) shows that feature $x$ is more important than
feature $y$ because the corresponding red circle for $x$ is higher than the
circle for $y$. At the same time, a part of the interval of $x$ for
$\varepsilon=0.15$ is smaller than the interval of $y$. This implies that the
decision made by using only the precise values of Shapley values may be
incorrect. A similar case is depicted in the right picture of Fig. \ref{f:12}.
Many such cases can be provided, which illustrate that imprecise Shapley
values should be studied instead of precise values because intervals show all
possible Shapley values for every feature and can reduce mistakes related to
the incorrect explanation.%

\subsection{Numerical experiments with real data}

In order to illustrate the Imprecise SHAP, we investigate the model for data
sets from UCI Machine Learning Repository \cite{Lichman:2013}. We consider
datasets: Seeds (https://archive.ics.uci.edu/ml/datasets/seeds), $m=7$,
$n=210$, $C=3$; Glass Identification
(https://archive.ics.uci.edu/ml/datasets/glass+identification), $m=10$,
$n=214$, $C=7$; Ecoli (https://archive.ics.uci.edu/ml/datasets/ecoli), $m=8$,
$n=336$, $C=8$. More detailed information can be found from, respectively, the
data resources.

\begin{figure}
[ptb]
\begin{center}
\includegraphics[
height=2.353in,
width=2.4188in
]%
{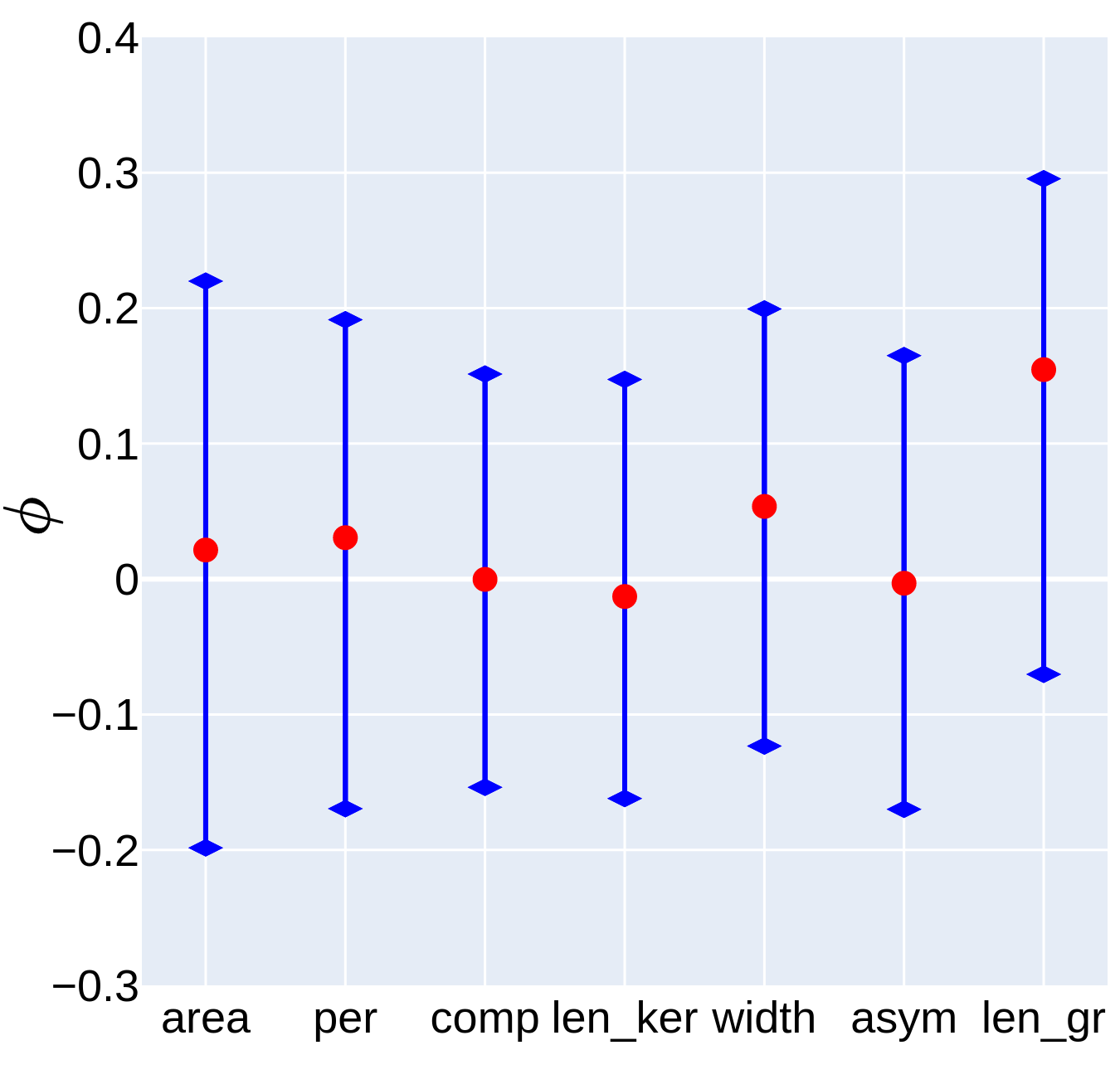}%
\caption{Intervals of Shapley values for $\varepsilon=0.15$ for the first
random point from the Seeds dataset}%
\label{f:seeds_6}%
\end{center}
\end{figure}

\begin{figure}
[ptb]
\begin{center}
\includegraphics[
height=2.4169in,
width=2.498in
]%
{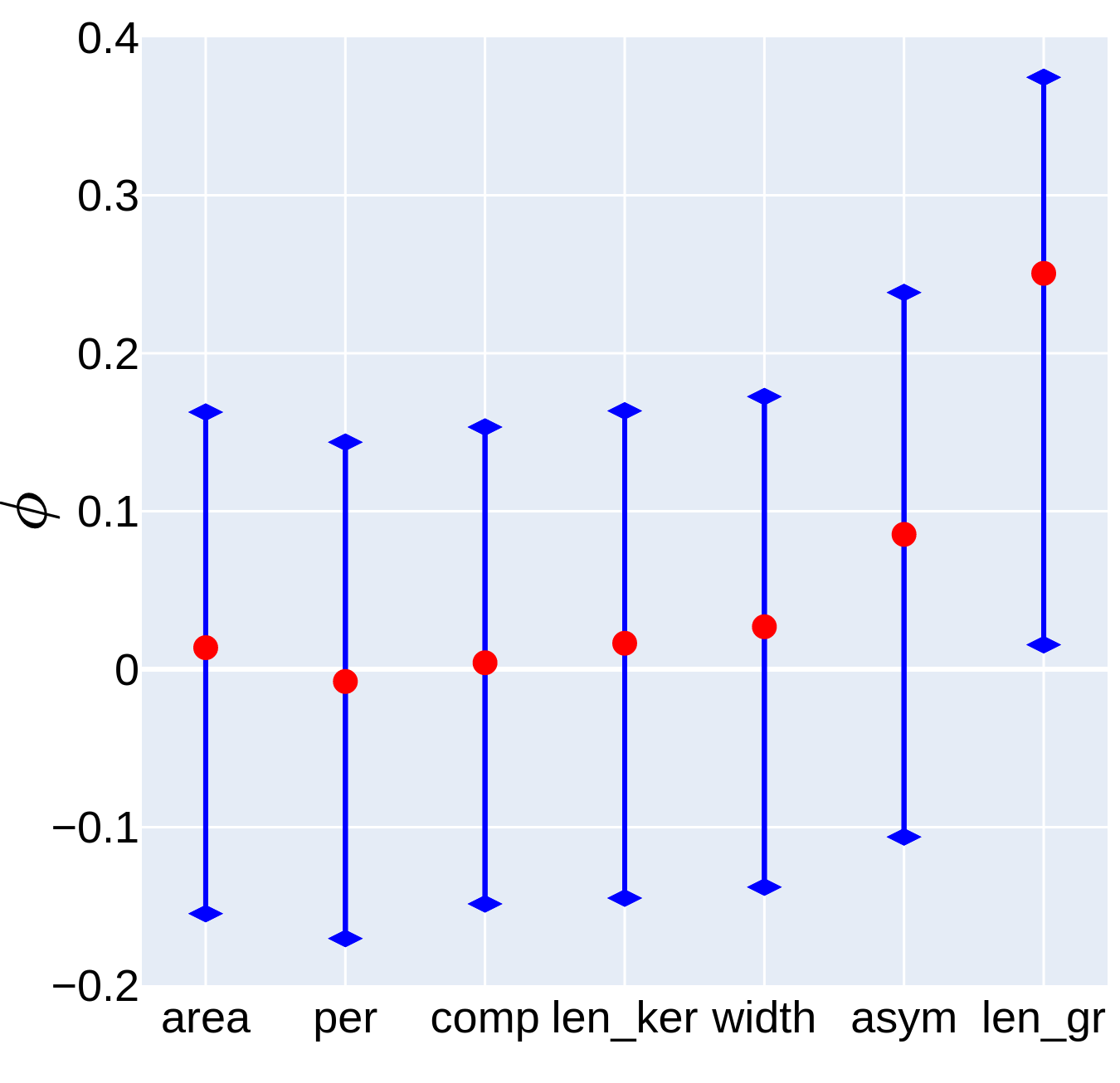}%
\caption{Intervals of Shapley values for $\varepsilon=0.15$ for the second
random point from the Seeds dataset}%
\label{f:seeds_max_46}%
\end{center}
\end{figure}

As an example, Fig. \ref{f:seeds_6} illustrates intervals of Shapley values
for random point $(16.7,14.7,0.9,6.4,3.5,2.5,5.6)$ from the Seeds dataset by
contamination parameter $\varepsilon=0.15$. Features are denoted as follows:
area (area), perimeter (per), compactness (comp), length of kernel (len\_ker),
width of kernel (width), asymmetry coefficient (asym), length of kernel groove
(len\_gr). Red circles, which are close to centers of intervals, show precise
Shapley values obtained for $\varepsilon=0$. It can be seen from Fig.
\ref{f:seeds_6} that feature \textquotedblleft len\_gr\textquotedblright%
\ provides the highest contribution to the corresponding prediction. In this
case, the relationship between intervals mainly coincides with the
relationship between precise values. However, every point in the intervals may
be regarded as a true contribution value, therefore, there is a chance that
feature \textquotedblleft len\_gr\textquotedblright\ does not maximally
contribute to the prediction. Another example for random point
$(18.1,16.1,0.88,6.06,3.56,3.62,6.0)$ by the same $\varepsilon$ is shown in
Fig. \ref{f:seeds_max_46} where the relationship between imprecise Shapley
values is more explicit.

Table \ref{t:original_SHAP_1} shows the original SHAP results
(https://github.com/slundberg/shap) and is provided for comparison purposes.
Its cell $(i,j)$ contains the value of the Shapley value for the $j$-th
feature under condition that the prediction is the $i$-the class probability.
Table \ref{t:original_SHAP_1} also shows that feature \textquotedblleft
len\_gr\textquotedblright\ has the largest contribution to the first and to
the second classes. This result correlates with the results shown in Fig.
\ref{f:seeds_max_46}.%

%

%

\begin{table}[tbp] \centering
\caption{The original SHAP results for the Seeds dataset and the second random point}%
\begin{tabular}
[c]{cccc}\hline
Features & Class 1 & Class 2 & Class 3\\\hline
area & $-0.019$ & $0.108$ & $-0.088$\\\hline
per & $-0.036$ & $0.109$ & $-0.073$\\\hline
comp & $0.005$ & $0.001$ & $-0.006$\\\hline
len\_ker & $-0.080$ & $0.128$ & $-0.048$\\\hline
width & $0.011$ & $0.04$ & $-0.051$\\\hline
asym & $-0.023$ & $0.018$ & $0.006$\\\hline
len\_gr & $-0.205$ & $0.254$ & $-0.049$\\\hline
\end{tabular}
\label{t:original_SHAP_1}%
\end{table}%

\begin{figure}
[ptb]
\begin{center}
\includegraphics[
height=2.389in,
width=2.48in
]%
{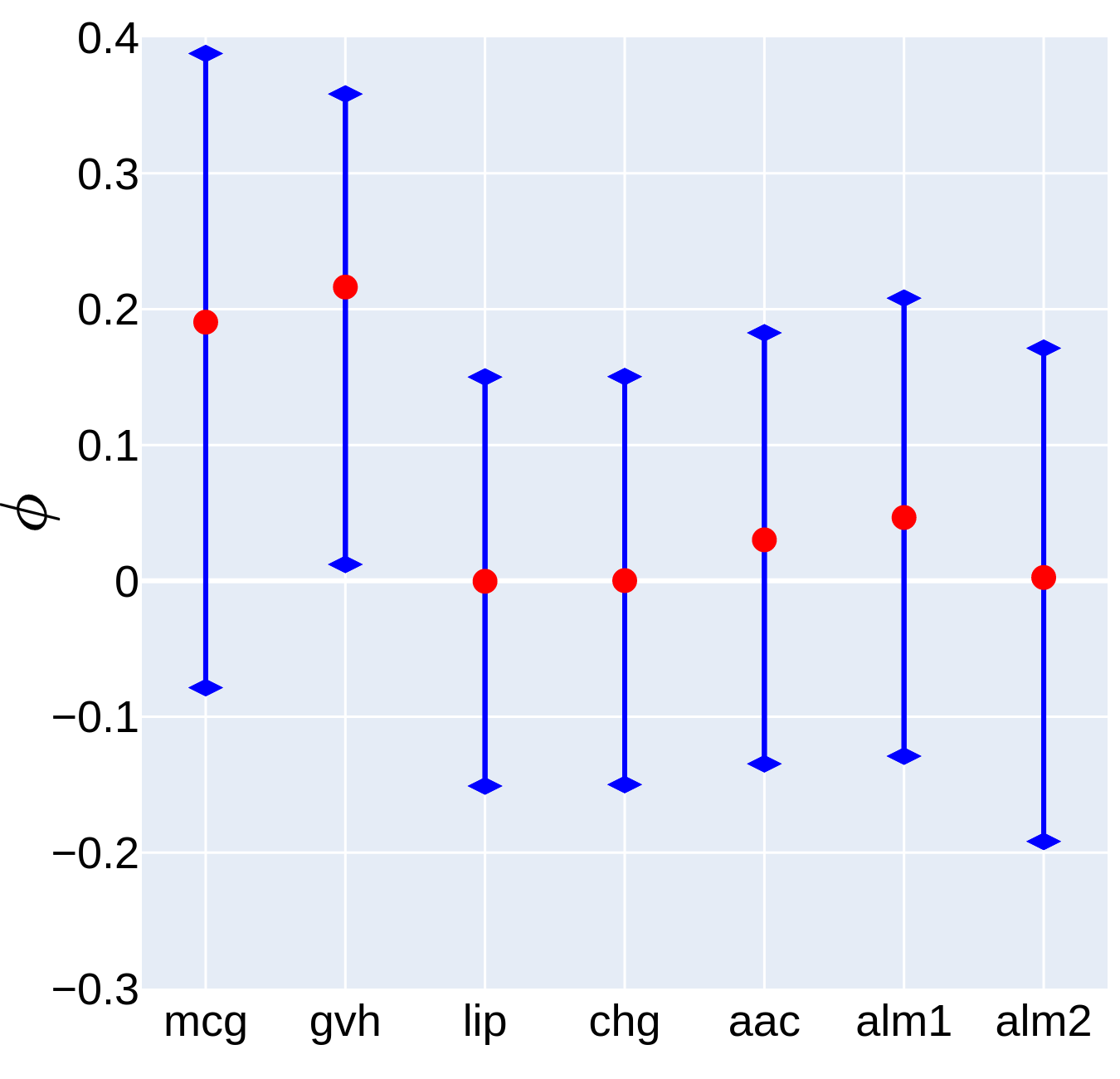}%
\caption{Intervals of Shapley values for $\varepsilon=0.15$ for the first
random point from the Ecoli dataset}%
\label{f:ecoli_4}%
\end{center}
\end{figure}

\begin{figure}
[ptb]
\begin{center}
\includegraphics[
height=2.453in,
width=2.5305in
]%
{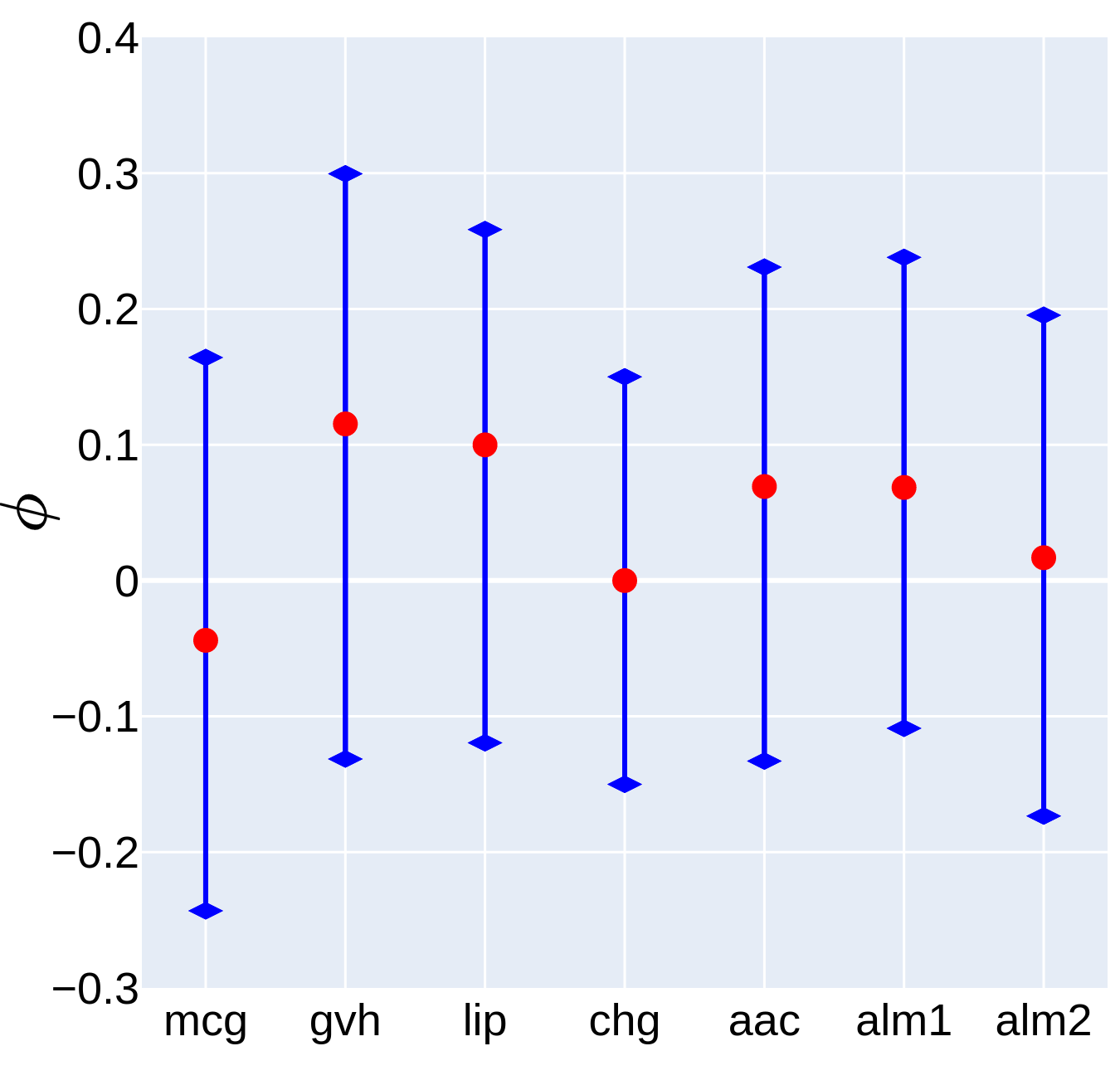}%
\caption{Intervals of Shapley values for $\varepsilon=0.15$ for the second
random point from the Ecoli dataset}%
\label{f:ecoli_13}%
\end{center}
\end{figure}

Fig. \ref{f:ecoli_4} shows intervals of Shapley values for random point
$(0.8,0.8,0.65,0.6,0.17,0.45,0.85)$ from the Ecoli dataset by contamination
parameter $\varepsilon=0.15$. Features are denoted in accordance with the data
resources as follows: mcg, gvh, lip, chg, aac, alm1, alm2. It can be seen from
Fig. \ref{f:ecoli_4} that two features \textquotedblleft mcg\textquotedblright%
\ and \textquotedblleft gvh\textquotedblright\ provide the highest
contribution to the corresponding prediction. However, it is difficult to
select a single feature among features \textquotedblleft mcg\textquotedblright%
\ and \textquotedblleft gvh\textquotedblright. From the one hand,
\textquotedblleft gvh\textquotedblright\ is more important if we assume that
$\varepsilon=0$. On the other hand, the corresponding interval of the Shapley
value included in interval of \textquotedblleft mcg\textquotedblright. This
example clearly illustrates that the use of imprecise Shapley values may
change our decision about the feature contributions. Another example for
random point $(0.5,0.79,0.8,0.6,0.8,0.58,0.3)$ by the same $\varepsilon$ is
shown in Fig. \ref{f:ecoli_13} where we have the same problem of the feature
selection among features \textquotedblleft gvh\textquotedblright\ and
\textquotedblleft lip\textquotedblright. Moreover, one can see from Fig.
\ref{f:ecoli_13} that features \textquotedblleft aac\textquotedblright\ and
\textquotedblleft alm1\textquotedblright\ can be also viewed as important
ones. This implies that imprecise Shapley values show a more correct
relationship of the feature contributions.%

%

\begin{figure}
[ptb]
\begin{center}
\includegraphics[
height=2.4881in,
width=2.5908in
]%
{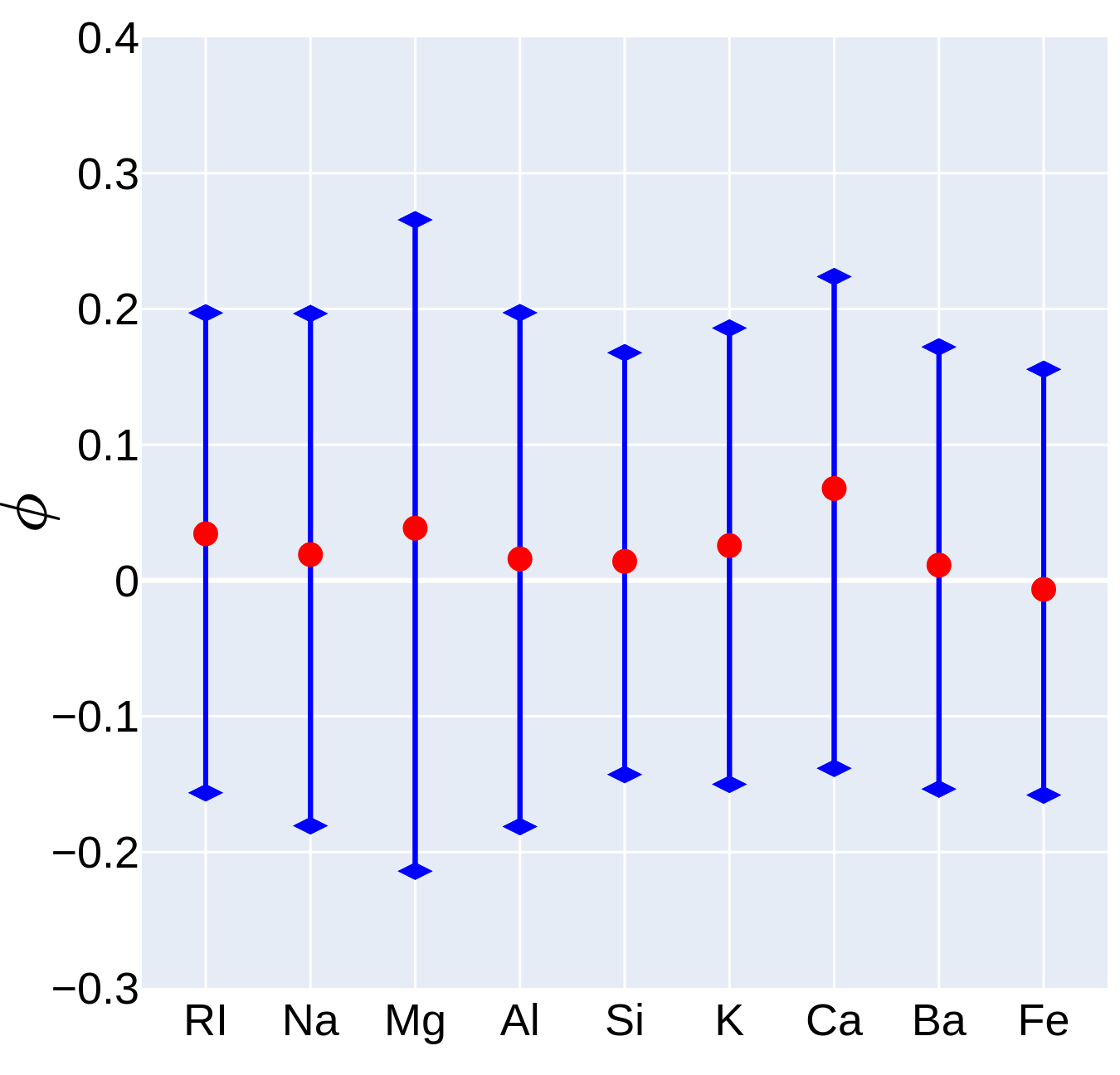}%
\caption{Intervals of Shapley values for $\varepsilon=0.15$ for the first
random point from the Glass Identification dataset}%
\label{f:lass_25}%
\end{center}
\end{figure}

\begin{figure}
[ptb]
\begin{center}
\includegraphics[
height=2.5206in,
width=2.6088in
]%
{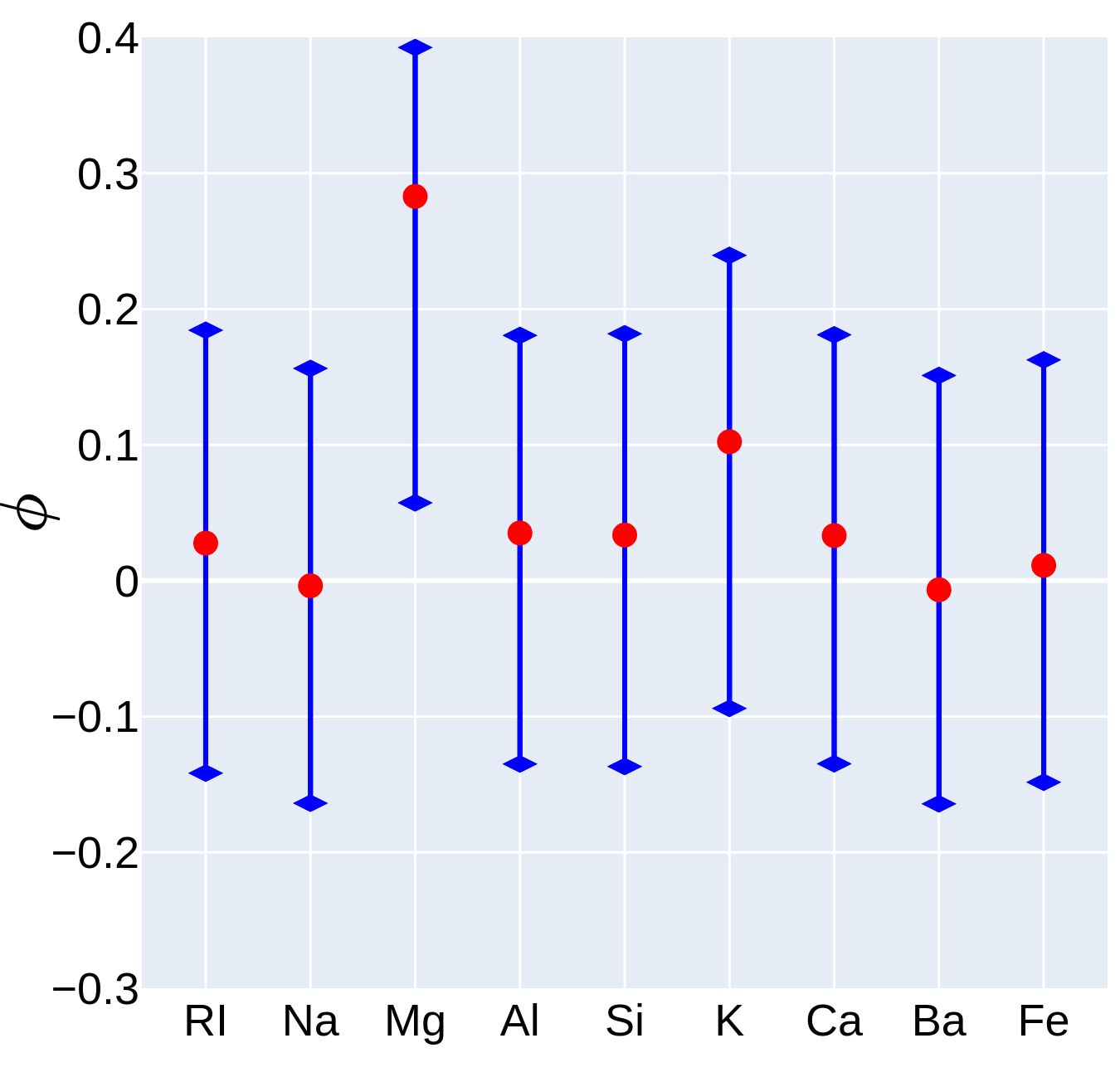}%
\caption{Intervals of Shapley values for $\varepsilon=0.15$ for the second
random point from the Glass Identification dataset}%
\label{f:lass_47}%
\end{center}
\end{figure}

Fig. \ref{f:lass_25} shows intervals of Shapley values for random point
$(1.53,16,2.15,0.56,72,3.6,13.1,0.07,0.005)$ from the Glass Identification
dataset by contamination parameter $\varepsilon=0.15$. Features are denoted in
accordance with the data resources as follows: RI, Na, Mg, Al, Si, K, Ca, Ba,
Fe. This numerical example is also demonstrative. One can see from Fig.
\ref{f:lass_25} that feature \textquotedblleft Ca\textquotedblright\ is the
most important under condition $\varepsilon=0$. However, the interval of this
feature is comparable with other intervals especially with the interval of
feature \textquotedblleft Mg\textquotedblright. In contrast to this numerical
example, imprecise Shapley values obtained for another random point
$(1.51,13,0.6,1.6,75,4.5,6.6,0.06,0.02)$ shown in Fig. \ref{f:lass_47}
strongly imply that feature \textquotedblleft Mg\textquotedblright\ is the
most important one.%

%

The above experiments have illustrated importance of the imprecise SHAP as a
method for taking into account possible aleatoric uncertainty of predictions
due to the limited amount of training data or other reasons. We could see from
the experiments that the precise explanation may differ from the imprecise
explanation. It should be noted that decision making based on intervals is
ambiguous. It depends on a selected decision strategy. In particular, the
well-known way for dealing with imprecise data is to use pessimistic or robust
strategy. In accordance with this strategy, if we are looking for the largest
values of contributions, then lower bounds of intervals are considered and
compared. If we are looking for unimportant features to remove them, then
upper bounds of intervals should be considered. For example, it follows from
the resulting imprecise Shapley values obtained for the second random point
from the Ecoli dataset, which are shown in Fig. \ref{f:ecoli_13}, the robust
strategy selects feature \textquotedblleft alm1\textquotedblright\ as the most
important one though the most important intervals by using precise Shapley
values ($\varepsilon=0$) provide three different important features:
\textquotedblleft gvh\textquotedblright, \textquotedblleft
lip\textquotedblright\ and \textquotedblleft aac\textquotedblright. The robust
strategy can be interpreted as an insurance against the worst case
\cite{Robert94}. Another \textquotedblleft extreme\textquotedblright\ strategy
is optimistic. It selects the upper bounds of intervals. The optimistic
strategy cannot be called robust. There are other strategies, for example, the
cautious strategy for which the most important feature is defined as
\begin{equation}
\arg\max_{i=1,...,m}\left(  \eta\phi_{i}^{L}+(1-\eta)\phi_{i}^{U}\right)  .
\end{equation}
Here $\eta$ is a cautious parameter taking values from $0$ to $1$. The case
$\eta=1$ corresponds to the robust decision strategy.

The choice of a certain strategy depends on a considered application and can
be regarded as a direction for further research.

\section{Conclusion}

A new approach to explanation of the black-box machine learning model
probabilistic predictions has been proposed, which aims to take into account
the imprecision of probabilities composing the predictions. It can be applied
to explanation of various machine learning models. In addition to
multiclassification models, we have to point out a class of survival models,
for example, the survival SVM \cite{Van_Belle-etal-2011}, random survival
forests \cite{Ibrahim-etal-2008,Wright-etal-2017}, survival neural networks
\cite{Faraggi-Simon-1995,Katzman-etal-2018}, where this approach can be
successfully used. Their predictions are survival functions which are usually
represented as stepwise functions due to a finite number of the event
observation times. Differences between steps of every survival function
produce a probability distribution. Due to a limited amount of data, we get
imprecise survival functions which have to be explained. A modification of
LIME called SurvLIME-KS \cite{Kovalev-Utkin-2020c} has been developed to solve
the survival model explanation problem under incomplete data. However,
SurvLIME-KS provides precise minimax coefficients of the LIME linear
regression which try to explain the worst case of the survival function from a
set of functions. Sometimes, it is useful to have interval-valued estimates,
which characterize the coefficients themselves as well as their uncertainty
(the interval width), instead of the robust values because it is difficult to
interpret the robustness itself. The proposed approach allows us to get
interval-valued Shapley values which have these properties.

It is important to point out that the presented results are general and do not
directly depend on the imprecise statistical model which describes imprecision
of predictions. At the same time, a specific implementation of the approach
requires choosing an imprecise model. In particular, bounds for $\pi_{i}$,
$\tau_{i}$, $\alpha_{i}$, $i=1,...,C-1$, are defined by (\ref{SHAP_prob_52})
which is given for the imprecise $\varepsilon$-contamination model. The
investigation how different imprecise models impact on the explanation results
can be regarded as a direction for further research. The same concerns with
the Kolmogorov-Smirnov distance. The approach adaptation to different
probability distribution distances is another direction for research.

A direction for further research is to study how to combine the obtained
intervals explaining predictions on parts of intervals to reduce the SHAP
complexity. This idea is mainly based on using an ensemble of random SHAPs
\cite{Utkin-Konstantinov-21} where the ensemble consists of many SHAPs such
that every SHAP explains only a part of features.

\section*{Acknowledgement}

This work is supported by the Russian Science Foundation under grant 21-11-00116.

\section*{Appendix}

\textbf{Proof of Proposition \ref{prop:imp_SHAP_1}}: Suppose that the sum
$\sum_{i=1}^{m}\phi_{i}$ is precise and equals to $t\in\lbrack D^{L},D^{U}]$.
First, we consider the upper bound $\ \tilde{\phi}_{k}^{U}$ for $\phi_{k}$.
Then the optimization problem is of the form:
\begin{equation}
\tilde{\phi}_{k}^{U}=\max\phi_{k},
\end{equation}
subject to
\begin{equation}
\sum_{i=1}^{m}\phi_{i}=t,
\end{equation}%
\begin{equation}
\phi_{i}^{L}\leq\phi_{i}\leq\phi_{i}^{U},~\phi_{i}\geq0,~i=1,...,m.
\end{equation}

Let us write the dual problem. It is of the form:
\begin{equation}
\tilde{\phi}_{k}^{U}=\min\left(  t\cdot w_{0}+\sum_{i=1}^{m}\phi_{i}^{U}\cdot
v_{i}-\sum_{i=1}^{m}\phi_{i}^{L}\cdot w_{i}\right)  , \label{SHAP_prob_31}%
\end{equation}
subject to
\begin{equation}
w_{0}+(v_{i}-w_{i})\geq\mathbf{1}_{k}(i),\ i=1,...,m, \label{SHAP_prob_32}%
\end{equation}%
\begin{equation}
v_{i}\geq0,\ w_{i}\geq0,\ i=1,...,m. \label{SHAP_prob_33}%
\end{equation}

Here $w_{0}\in\mathbb{R}$, $v_{i}\in\mathbb{R}_{+}$, $w_{i}\in\mathbb{R}_{+}$
are optimization variables; $\mathbf{1}_{k}(i)$ is the indicator function
taking value $1$ if $k=i$. The problem has $2m+1$ variables and $3m$
constraints. It is well-known from the linear programming theory that $2m+1$
constraints among all constraints are equalities. It is simply to prove that
either $v_{i}=0$ or $w_{i}=0$. Suppose that the $k$-th constraint is one of
the equalities. Suppose also that $v_{k}=1$ and $w_{k}=0$ (only for the $k$-th
constraint). To minimize the objective function, $w_{0}$ should be as small as
possible. The assumption $\phi_{i}^{L}\geq0$ for all $i$ is used here. It
follows from the $k$-th constraint that $w_{0}=0$. Then all variables $v_{i}$,
$w_{i}$ are zero. Substituting the values into objective function, we get the
first solution $\tilde{\phi}_{k}^{U}=\phi_{k}^{U}$. Another solution when the
$k$-th constraint is one of the equalities is obtained if $w_{0}=1$, $v_{k}%
=0$, $w_{k}=1$. In this case, $v_{i}=0$ and $w_{i}=1$ for all $i\neq k$.
Hence, there holds
\begin{equation}
\tilde{\phi}_{k}^{U}=t-\sum_{i=1}^{m}\phi_{i}^{L}.
\end{equation}

Other combinations of equalities lead to a larger objective function. In sum,
we get the solution of problem (\ref{SHAP_prob_31})-(\ref{SHAP_prob_33}):
\begin{equation}
\tilde{\phi}_{k}^{U}=\min\left(  \phi_{k}^{U},t-\sum_{i=1}^{m}\phi_{i}%
^{L}\right)  .
\end{equation}

The solution is valid for all $t$ from interval $[D^{L},D^{U}]$. This implies
that the smallest $\tilde{\phi}_{k}^{U}$ is achieved when $t=D^{L}$, and we
get (\ref{SHAP_prob_28}). The same approach can be used to derive the lower
bound $\tilde{\phi}_{i}^{L}$ which is given in (\ref{SHAP_prob_29}) as was to
be proved.

\textbf{Proof of Proposition \ref{prop:Low_Dist}}: Introduce a new variable
$B=\sup_{i=1,..,C-1}\left\vert \pi_{i}-\alpha_{i}\right\vert $. Then problem
(\ref{SHAP_prob_55}) becomes%
\begin{equation}
\min_{\pi,\tau,\alpha}\left(  B-\sup_{i=1,..,C-1}\left\vert \tau_{i}%
-\alpha_{i}\right\vert \right)  ,
\end{equation}
subject to (\ref{SHAP_prob_54}) and (\ref{SHAP_prob_58}).

Constraints (\ref{SHAP_prob_58}) mean that $\pi_{i}$ and $\alpha_{i}$ are
elements of the cumulative distribution functions. Suppose that the largest
value of $\left\vert \tau_{i}-\alpha_{i}\right\vert $ is achieved for some
index $k$, i.e.,
\begin{equation}
k=\arg\max_{i}\left\vert \tau_{i}-\alpha_{i}\right\vert .
\end{equation}
Then we can rewrite the objective function for the given $k$ as
\begin{equation}
~\min_{\pi,\tau,\alpha}\left(  B-\left\vert \tau_{k}-\alpha_{k}\right\vert
\right)  .
\end{equation}

To achieve the minimum of the objective function, the term $\left\vert
\tau_{k}-\alpha_{k}\right\vert $ has to be maximized. Since, the cumulative
distribution $\tau$ does not depend on $\pi$ and $\alpha$, then the maximum of
$\left\vert \tau_{k}-\alpha_{k}\right\vert $ is reduced to two cases:%
\begin{align*}
\max_{\tau}\tau_{k}-\alpha_{k}  &  =\tau_{k}^{U}-\alpha_{k},\\
\alpha_{k}-\min_{\tau}\tau_{k}  &  =\alpha_{k}-\tau_{k}^{L}.
\end{align*}

The first case requires condition $\alpha_{k}\leq\tau_{k}^{U}$. Condition
$\alpha_{k}\geq\tau_{k}^{L}$ is for the second case. In sum, we get $C-1$
linear programming problems (\ref{SHAP_prob_60}) and $C-1$ problems
(\ref{SHAP_prob_61}) with the same constraints (\ref{SHAP_prob_56}%
)-(\ref{SHAP_prob_58}) and different constraints (\ref{SHAP_prob_59}). It is
obvious that the final solution $L$ of (\ref{SHAP_prob_55}) is determined by
comparison of all optimal terms $\alpha_{k}-\tau_{k}^{U}$ from
(\ref{SHAP_prob_60}) and optimal terms $\tau_{k}^{L}-\alpha_{k}$ from
(\ref{SHAP_prob_61}).

\textbf{Proof of Proposition \ref{prop:Upp_Dist}}: The proof is similar to the
proof of Proposition \ref{prop:Low_Dist}.

\textbf{Proof of Corollary \ref{cor:SHAP1}}: Solutions (\ref{SHAP_prob_83})
and \ref{SHAP_prob_84}) directly follows from the proof of Propositions
\ref{prop:Low_Dist}-\ref{prop:Upp_Dist} and from considering all variants of
signs of $\pi_{1}-\alpha_{1}$ and $\tau_{1}-\alpha_{1}$.

\textbf{Proof of Proposition \ref{prop:general_bounds}}: The lower bound is
simply derived by introducing a new variable $B=\max_{i=1,...,C-1}\left\vert
\pi_{i}-\alpha_{i}\right\vert $. The upper bound $D^{U}$ can be derived
similarly to problems (\ref{SHAP_prob_60})-(\ref{SHAP_prob_59}) or
(\ref{SHAP_prob_70})-(\ref{SHAP_prob_79}). Let
\begin{equation}
k=\arg\max_{i}\left\vert \tau_{i}-\alpha_{i}\right\vert .
\end{equation}
Introduce
\[
D^{U}(k)=\max_{\pi,\alpha}\left\vert \pi_{k}-\alpha_{k}\right\vert .
\]
Hence, we get two problems. The first one is
\[
D_{1}^{U}(k)=\max_{\pi,\alpha}\left(  \pi_{k}-\alpha_{k}\right)  ,
\]
subject to (\ref{SHAP_prob_56}) and $\pi_{k}\geq\alpha_{k}$.

The second problem is%
\[
D_{2}^{U}(k)=\max_{\pi,\alpha}\left(  \alpha_{k}-\pi_{k}\right)  ,
\]
subject to (\ref{SHAP_prob_56}) and $\alpha_{k}\geq\pi_{k}$.

Solutions are trivial and have the form:
\[
D^{U}(k)=\max\left\{
\begin{array}
[c]{cc}%
\pi_{k}^{U}-\alpha_{k}^{L}, & \text{if }\pi_{k}^{U}\geq\alpha_{k}^{L},\\
\alpha_{k}^{U}-\pi_{k}^{L} & \text{if }\pi_{k}^{L}<\alpha_{k}^{U}.
\end{array}
\right.
\]

In sum, we get the upper bound as
\[
D^{U}=\max_{k=1,...,C-1}D^{U}(k),
\]
as was to be proved.

\bibliographystyle{plain}
\bibliography{Boosting,Classif_bib,Deep_Forest,Explain,Explain_med,Imprbib,Lasso,Math_bib,MYBIB,MYUSE,Robots,Survival_analysis}

\begin{thebibliography}{10}
\bibitem{Pearson-1900}
K.~Pearson.
\newblock On the criterion that a given system of eviations from the probable
  in the case of correlated system of variables is such that it can be
  reasonable supposed to have arisen from random sampling.
\newblock {\em Philosophical Magazine}, 50:157--172, 1900.
\end{thebibliography}

\end{document}